# Enhancing Cancer Diagnosis with Explainable & Trustworthy Deep Learning Models

Badaru I. Olumuyiwa, Author, The Anh Han and Zia U. Shamszaman.

*Abstract*— **This research presents an innovative approach to cancer diagnosis and prediction using explainable Artificial Intelligence (XAI) and deep learning techniques. With cancer causing nearly 10 million deaths globally in 2020, early and accurate diagnosis is crucial. Traditional methods often face challenges in cost, accuracy, and efficiency. Our study develops an AI model that provides precise outcomes and clear insights into its decision-making process, addressing the "black box" problem of deep learning models. By employing XAI techniques, we enhance interpretability and transparency, building trust among healthcare professionals and patients. Our approach leverages neural networks to analyse extensive datasets, identifying patterns for cancer detection. This model has the potential to revolutionise diagnosis by improving accuracy, accessibility, and clarity in medical decision-making, possibly leading to earlier detection and more personalised treatment strategies. Furthermore, it could democratise access to high-quality diagnostics, particularly in resource-limited settings, contributing to global health equity. The model's applications extend beyond cancer diagnosis, potentially transforming various aspects of medical decision-making and saving millions of lives worldwide.**

## I. Introduction

Cancer is one of the leading causes of death globally, responsible for nearly 10 million deaths in 2020 [1]. Early detection and accurate diagnosis are crucial for improving patient outcomes and survival rates. However, traditional diagnostic methods often face challenges in cost, accuracy, and efficiency [3]. Whilst cancer has been a formidable challenge to human health throughout history, our understanding and treatment methods have evolved significantly. The discovery of X-rays in the late 1800s marked a turning point, paving the way for modern medical oncology [2].

Today, breast, lung, colorectal, prostate, and skin cancers are the most common types. Various factors influence cancer rates, including age, gender, race, ethnicity, lifestyle choices, environment, genetics, and healthcare access [1]. Traditional cancer detection and treatment methods, such as medical imaging, chemotherapy, and surgery, whilst advanced, still have drawbacks in terms of cost, accuracy, and side effects [3].

Artificial intelligence (AI) has the potential to address these issues and revolutionise healthcare, especially in cancer treatment. AI can analyse vast amounts of data, identify patterns, and offer insights to inform decision-making, potentially reducing human error, bias, and costs whilst improving healthcare efficiency [4]. Deep learning, in particular, has shown remarkable accuracy in areas like genetic analysis, image analysis, and clinical decision support for cancer diagnosis and prognosis [5].

However, using AI in healthcare comes with challenges, particularly regarding the lack of transparency and interpretability in many AI models [6][7]. This "black box" nature can undermine trust and hinder adoption in clinical settings.

Our research project seeks to address this issue by developing a reliable, interpretable AI model for cancer diagnosis and prediction using state-of-the-art deep learning techniques and explainable AI (XAI) methods. By enhancing understanding of AI predictions, we aim to improve patient and healthcare provider trust in AI-assisted clinical decision-making [8][9], potentially transforming cancer diagnosis and prediction, and ultimately saving lives and improving patient outcomes.

## II. Methodology: System Design, Architecture and Implementation

In this section, we describe the system design, architecture, and implementation of the proposed AI system, which aims to achieve predictability, explainability, and interpretability

### A. System Requirements

We have conducted extensive research to identify the system requirements, which include:

Dataset: Several sources of cancer-related data were evaluated in order to identify a relevant dataset from Kaggle [10], which provides a large number of potentially helpful

This paragraph of the first footnote will contain the date on which you submitted your paper for review. It will also contain support information, including sponsor and financial support acknowledgment. For example, “This work was supported in part by the U.S. Department of Commerce under Grant BS123456.”

The next few paragraphs should contain the authors’ current affiliations, including current address and e-mail. For example, F. A. Author is with the National Institute of Standards and Technology, Boulder, CO 80305 USA (e-mail: author@ boulder.nist.gov).

S. B. Author, Jr., was with Rice University, Houston, TX 77005 USA. He is now with the Department of Physics, Colorado State University, Fort Collins, CO 80523 USA (e-mail: author@lamar.colostate.edu).

T. C. Author is with the Electrical Engineering Department, University of Colorado, Boulder, CO 80309 USA, on leave from the National Research Institute for Metals, Tsukuba, Japan (e-mail: author@nrim.go.jp).

datasets. Since data is essential to a trained model's ability to make predictions and produce outcomes, the features required for the proposed AI system's performance were sought after [11]. According to [12], a less biased dataset was sought to guarantee fairness during training and enhance the model's robustness and generalization.

The data's privacy and security concerns were also taken into account. The implementation of robust data privacy and security measures is not only an ethical imperative but also a legal requirement in many jurisdictions.

1. GDPR Compliance

The General Data Protection Regulation (GDPR) sets a high standard for data protection in the European Union and has become a global benchmark. Our system adheres to key GDPR principles, including:

- Data Minimization: Only the data necessary for cancer diagnosis are collected and processed, avoiding extraneous information.
- Purpose Limitation: Patient data is used exclusively for the stated purpose of cancer diagnosis and related research.
- Storage Limitation: We implement strict data retention policies, securely deleting data when it's no longer needed.
- Transparency: Clear communication with patients about data usage and AI involvement in diagnosis.
- Right to Erasure: Implementing mechanisms for patients to request deletion of their data, where applicable.

2. HIPAA Compliance

For deployments in the United States, our system is designed to meet the stringent requirements of the Health Insurance Portability and Accountability Act (HIPAA), including:

- Privacy Rule: Ensuring proper use and disclosure of protected health information (PHI).
- Security Rule: Implementing appropriate administrative, physical, and technical safeguards.
- Breach Notification Rule: Establishing protocols for timely notification in case of data breaches.

3. Technical Measures for Data Privacy and Security

Encryption Methods: To protect data both at rest and in transit, state-of-the-art encryption techniques were employed:

- Data Store: Utilization of AES-256 encryption for stored data. Implementation of secure key management practices, including regular key rotation.

- Data in Transit: Employment of TLS 1.3 protocols for all data transfers. Use of secure APIs with proper authentication mechanisms.

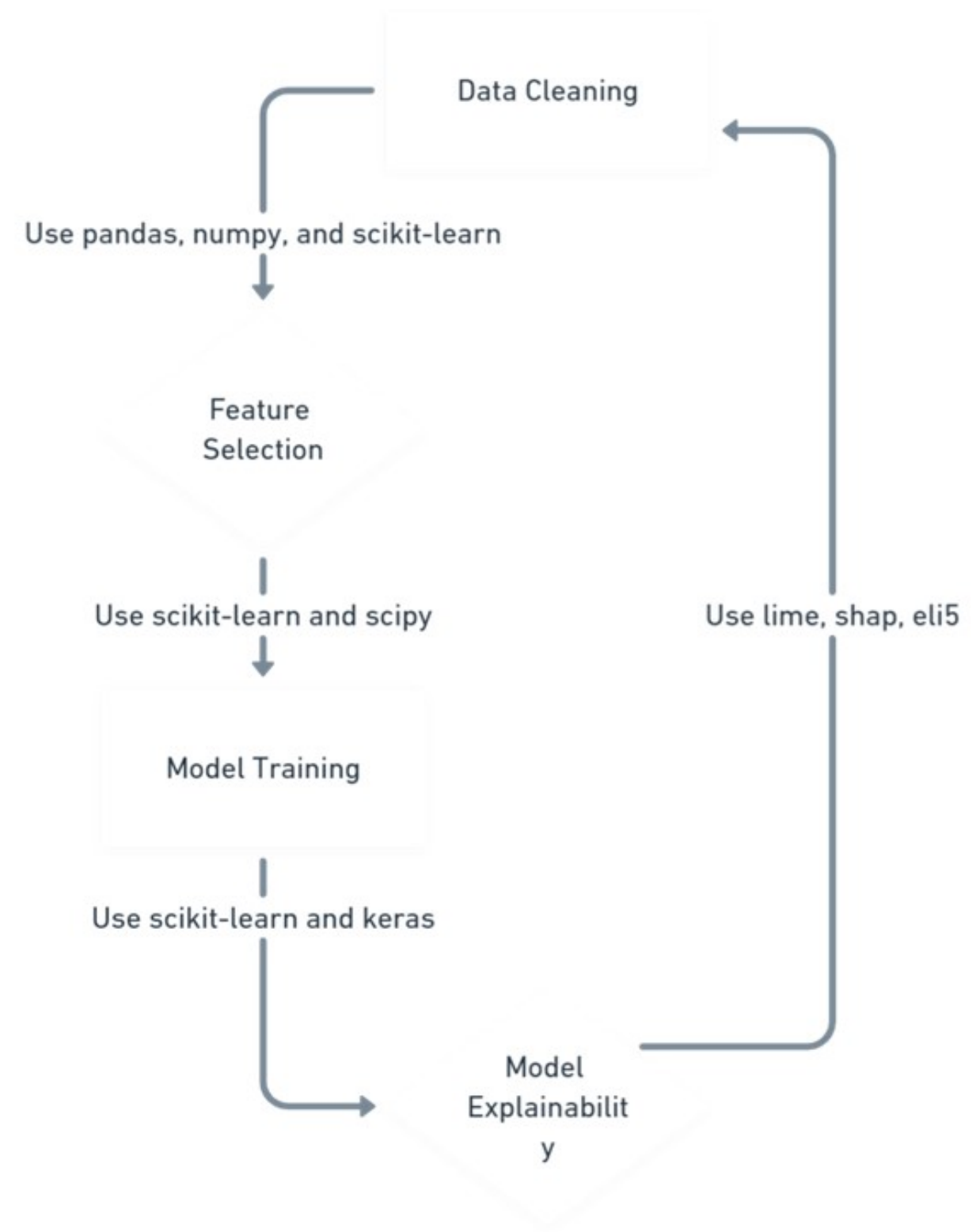

Fig. 1. Methodology Flowchart.

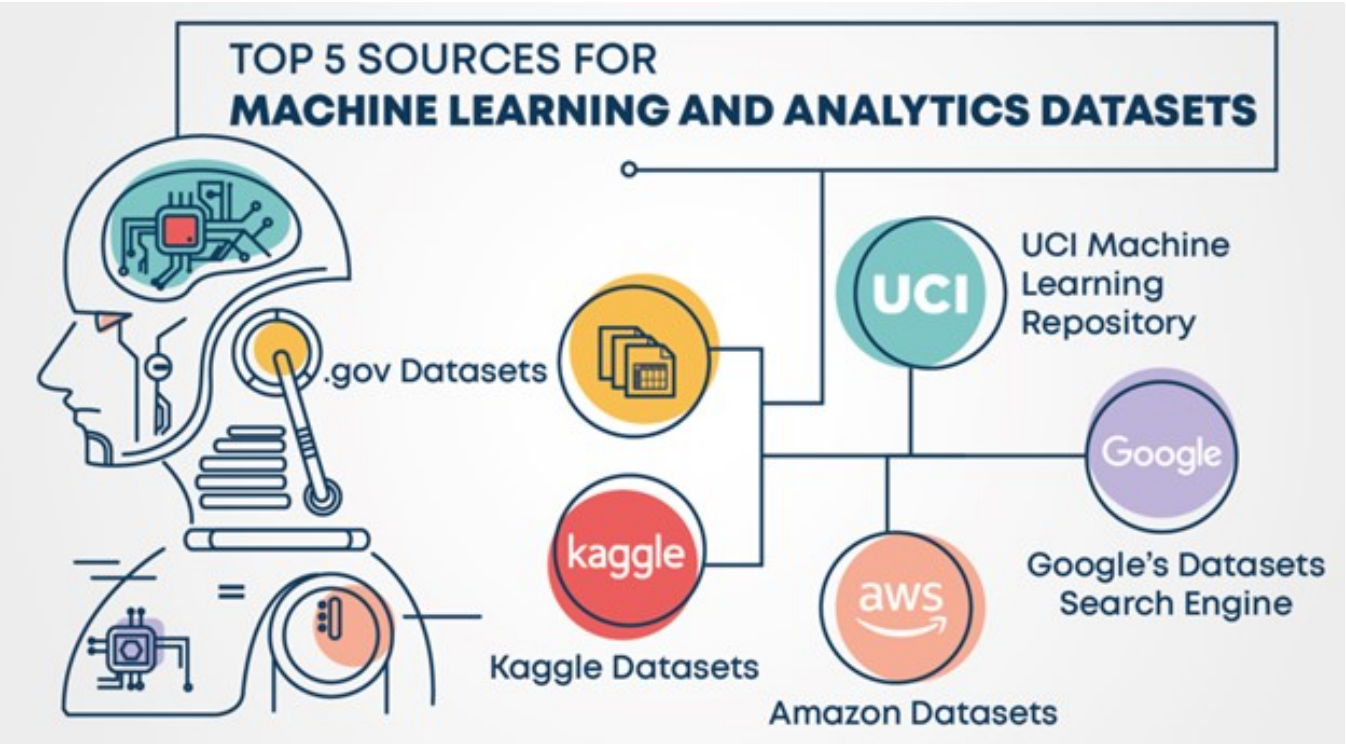

Fig. 2. Top 5 Sources for Datasets [13].

### *B. Data Pre-processing and Feature Engineering*

Pre-processing and feature engineering of the data are essential for ensuring that it is error-free and algorithm-compatible during dataset analysis and training [11]. New elements that might improve the training process were created, and the ones that are unnecessary for the model training were eliminated. Along with filling up the dataset's missing values, the data types of the features were examined, and some columns underwent categorization. To improve the model's performance when trained on the data, the dataset's normality was also examined and adjusted. The performance of the model depends on this prerequisite.

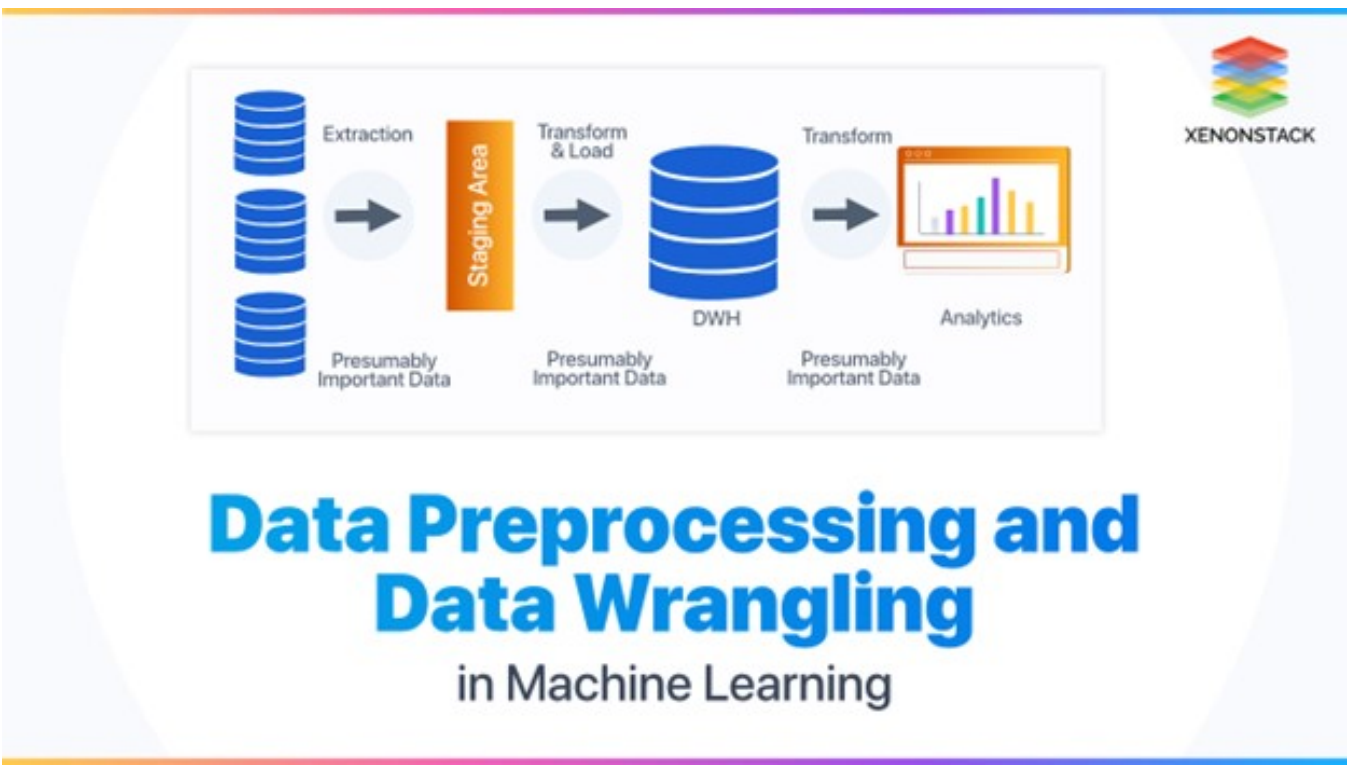

Fig. 3. Data Preprocessing and Data Wrangling [14].

### *C. Deep Learning Model*

The algorithm used to handle the pre-processed data is referred to in this requirement [11]. Similar to how the human brain functions, it matches patterns in neural networks and learns from them. It is made up of learning neurons as well as additional layers of neurons—also referred to as the model's hidden layers—beyond the input and output layer. Deep learning algorithms come in a variety of forms, and some of them will be applied in this project.

The relatively limited size of our dataset, comprising 569 cases, presents a substantial risk of overfitting. This phenomenon occurs when a model excessively adapts to the training data, including its inherent noise and fluctuations, resulting in poor generalization performance on novel, unseen data.

To mitigate this risk, we implemented the following techniques:

- Data Augmentation: This represents a fundamental approach to artificial dataset expansion, serving as a critical regularisation technique in deep learning applications.This approach involves the creation of synthetic examples through the application of transformations to existing data points. In the context of our study, this could encompass minor rotations or scaling of cell images. Data augmentation effectively expands the training set, potentially enhancing the model's ability to generalize.

  In the context of image processing tasks, these transformations typically manifest as geometric manipulations (e.g., rotations, translations, and scaling operations) or photometric adjustments (including variations in brightness, contrast, and colour balance). The theoretical underpinning of this approach lies in the assumption that such transformations preserve class-relevant information whilst introducing beneficial variability into the training process. The efficacy of this technique stems from its ability to:
  - Reduce overfitting by exposing the model to a broader range of valid input variations
  - Enhance invariance to specific transformations relevant to the task domain
  - Mitigate class imbalance issues through targeted augmentation of underrepresented classes
- Dropout Layers: This technique involves the random "dropping out" or deactivation of a proportion of neurons during the training process. Dropout serves to prevent co-adaptation of feature detectors and has been demonstrated to improve model generalization. The optimal dropout rate will be determined through empirical testing.

  Dropout represents a sophisticated regularisation technique that has garnered significant attention in the deep learning community since its introduction by [31]. This approach involves the stochastic omission of neural units during the training phase, effectively creating an ensemble of subnetworks within the primary architecture.

  The mathematical framework underlying dropout can be expressed as:

  $y = f(Wx) \odot m,$

  where

  m ~ Bernoulli(p)

  Here, m represents a binary mask sampled from a Bernoulli distribution with parameter p, typically set between 0.2 and 0.5. This formulation results in several theoretical advantages:
  - Prevention of Feature Co-adaptation: By randomly deactivating neurons, dropout disrupts the formation of excessive co-dependencies between neural units
  - Implicit Ensemble Learning: The technique effectively trains an ensemble of 2^n thinned networks, where n represents the number of units subject to dropout
  - Reduced Overfitting: The stochastic nature of dropout introduces beneficial noise into the training process, enhancing generalisation capabilities

- Regularization: The incorporation of L1 or L2 regularization terms in the model's objective function can effectively penalize overly complex models. This approach encourages the development of simpler, more generalizable solutions by adding a cost associated with large weights in the model.
- Early Stopping: Early stopping constitutes a pragmatic approach to optimisation control in neural network training. This methodology involves the continuous monitoring of model performance on a validation dataset, with training termination occurring when generalisation performance begins to deteriorate.
  The theoretical justification for early stopping stems from the observation that neural networks typically exhibit distinct phases during training:
  - Initial Learning Phase: Characterised by rapid improvement in both training and validation

performance
  - Optimal Generalisation Point: Where validation performance reaches its peak
  - Overfitting Phase: Marked by continued improvement in training performance but degradation in validation metrics

Implementation typically involves:
  - Performance Monitoring: Regular evaluation of model performance on a held-out validation set
  - Stopping Criterion: Definition of specific conditions that trigger training termination
  - Model Selection: Retention of the model state that achieved optimal validation performance

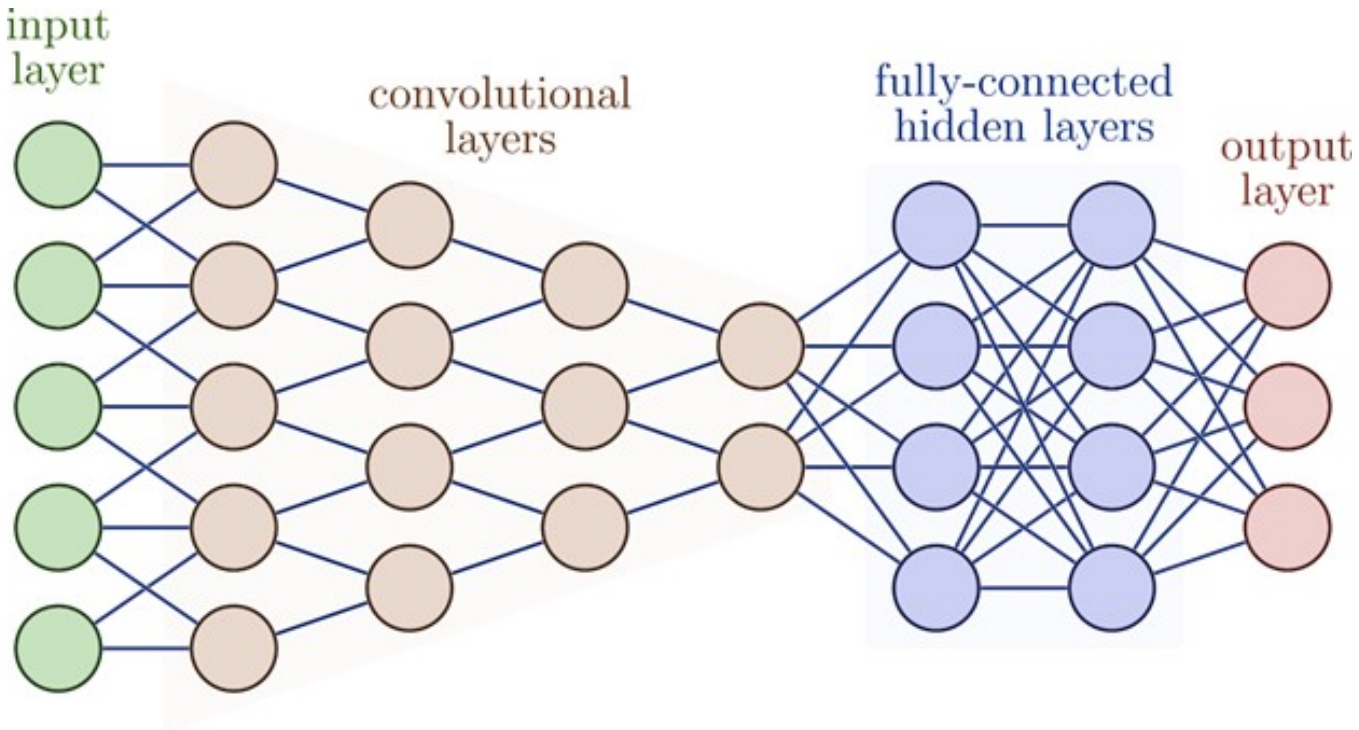

Fig. 4. Convolutional Neural Network Diagram [15].

### D. *Explainable AI Frameworks/Modules*

An essential prerequisite for analysing the inner workings of the black box model is Explainable AI Frameworks/Modules [8]. It will be applied to interpret and elucidate the reasons behind the predictions made by the different models that will be employed. This regulation will resolve these kinds of problems. For this research to be successful, cutting-edge XAI frameworks are required [12].

1. SHAP (SHapley Additive exPlanations)

SHAP, rooted in cooperative game theory, quantifies the contribution of each feature to the prediction for individual instances. This method is based on Shapley values, a concept from game theory that fairly distributes payout among players in a cooperative game.

a. Methodology

SHAP evaluates all possible combinations of features and calculates the marginal contribution of each feature to the difference between the actual prediction and the average prediction. This comprehensive approach ensures a robust and theoretically grounded assessment of feature importance. To properly understand this, let's delve into its inner workings:

The Mathematics Behind SHAP

In SHAP's framework, we treat feature attribution as a cooperative game where:

- The "players" are individual features in our dataset
- The "game" is the prediction task
- The "payout" is the difference between the model's prediction and the average prediction

The Shapley value for a feature i is calculated as:

$$\varphi_i = \Sigma\,(|S|!\,(n - |S| - 1)!/n!)\,[fx(S \cup \{i\}) - fx(S)] \quad (1)$$

Where:

- S represents all possible subsets of features excluding feature i
- n is the total number of features
- fx(S) is the model's prediction with only the features in subset S

The Coalitional Computation Process

- SHAP examines every possible coalition (combination) of features
- For each feature, it calculates:
  - The model's prediction with the feature present
  - The model's prediction with the feature absent
  - The weighted difference between these predictions
- This process is repeated across all possible feature combinations

b. Contributions to Model Interpretability

- Dual-level Explanations: SHAP provides both global and local interpretability. At the global level, it offers insights into overall feature importance across the dataset. At the local level, it elucidates the impact of features on individual predictions.
- Theoretical Soundness: The method's foundation in game theory lends it a strong theoretical basis, ensuring consistency and reliability in its explanations.
- Non-linear Relationship Handling: SHAP effectively captures and explains complex, non-linear relationships between features and model outputs.

2. LIME (Local Interpretable Model-agnostic Explanations)

LIME focuses on creating locally interpretable models to explain individual predictions of black-box models. This approach aims to approximate the behavior of complex models in local regions around specific predictions.

a. Methodology

LIME operates by perturbing the input data and observing the corresponding changes in model predictions. It then fits a simple, interpretable model (e.g., linear regression) to this local region, providing an approximation of the complex model's behavior in the vicinity of the prediction of interest. Here's a detailed examination of its methodology:

The Technical Implementation

- Sampling and Perturbation:
  - Generate synthetic samples around the instance of interest
  - Apply small perturbations to feature values
  - Weight samples based on their proximity to the original instance

Local Model Fitting:
The algorithm fits an interpretable model g to minimise:

$$argmin\ g \in G\ L(f, g, \pi x) + \Omega(g) \quad (2)$$

Where:

- f is the complex model
- g is the interpretable model
- $\pi$x is the locality weighting kernel
- $\Omega$(g) is a complexity penalty

The Mathematical Framework

LIME employs an exponential kernel for similarity weighting:

$$\pi x(z) = exp(-D(x,z)^2/\sigma^2) \quad (3)$$

Where:

- x is the original instance
- z is the perturbed instance
- D is the distance function
- $\sigma$ is the kernel width parameter

b. Contributions to Model Interpretability

- Local Explanations: LIME excels in providing easily comprehensible explanations for individual predictions, enhancing understanding of specific model decisions.
- Model Agnosticism: As a model-agnostic approach, LIME can be applied to any black-box model, offering flexibility across various ML architectures.
- Intuitive Understanding: By approximating complex models with simpler, local models, LIME facilitates an intuitive grasp of model behavior in specific instances.

3. Eli5 Permutation Importance

Eli5 is a Python library that provides various tools for model interpretation, including Permutation Importance, which is a model-agnostic method for determining feature importance.

a. Methodology

Permutation Importance works by randomly shuffling the values of each feature and measuring the resulting decrease in model performance. The features that, when shuffled, cause the largest decrease in performance are considered the most important.

b. Contributions to Model Interpretability

- Global Feature Importance: Eli5's Permutation Importance provides a clear ranking of feature importance at the global level, helping to identify which features have the most significant impact on the model's predictions overall.
- Model Agnosticism: Like LIME, this method can be applied to any type of model, making it versatile across different ML architectures.
- Simplicity and Intuitiveness: The concept behind Permutation Importance is straightforward and easy to explain, making it accessible to non-technical stakeholders.
- Computational Efficiency: Compared to SHAP, Permutation Importance is generally less computationally intensive, especially for large datasets or complex models.

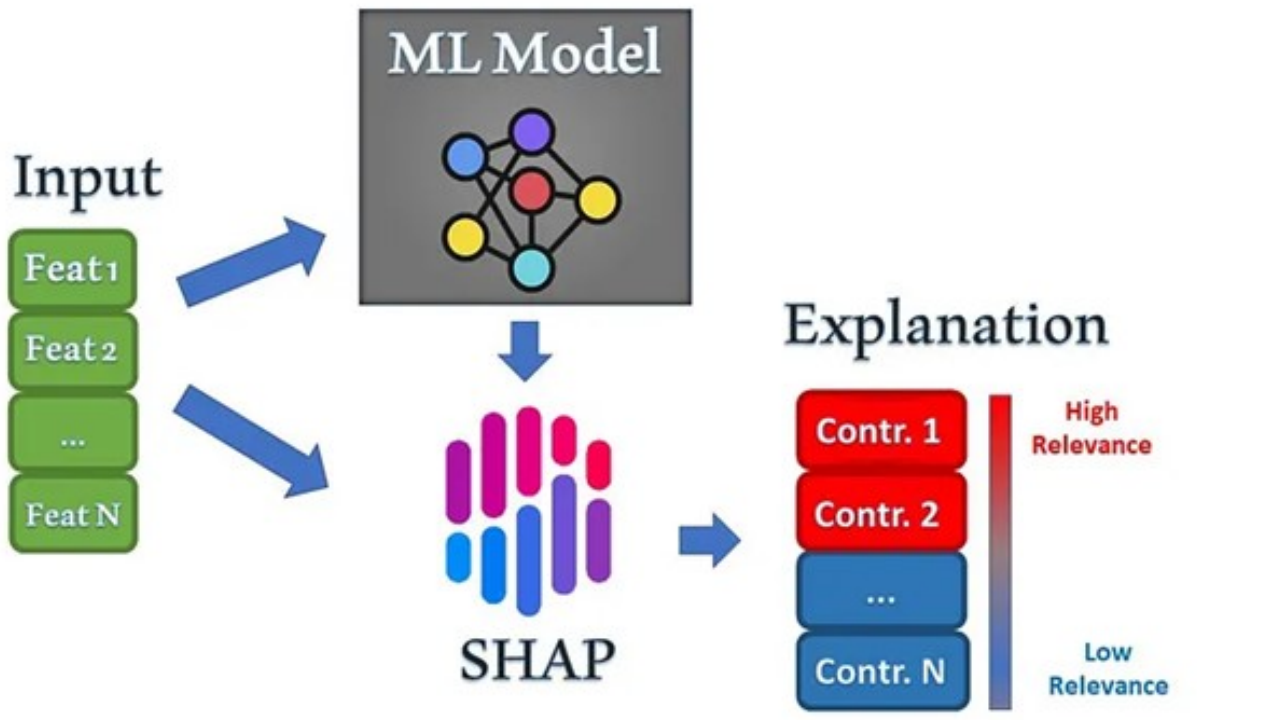

Fig. 5. Diagram of XAI Framework–[SHAP] for Model Explainability [16].

## *E. Data Visualization*

Data visualization is a crucial component of the model and XAI module results presentation, providing a graphical depiction of the entire outcome for effortless comprehension. Plots, histograms, heatmaps, and other visualizations [17] will be utilized to convey the insights gleaned from the datasets as well as the interpretability and model predictions.

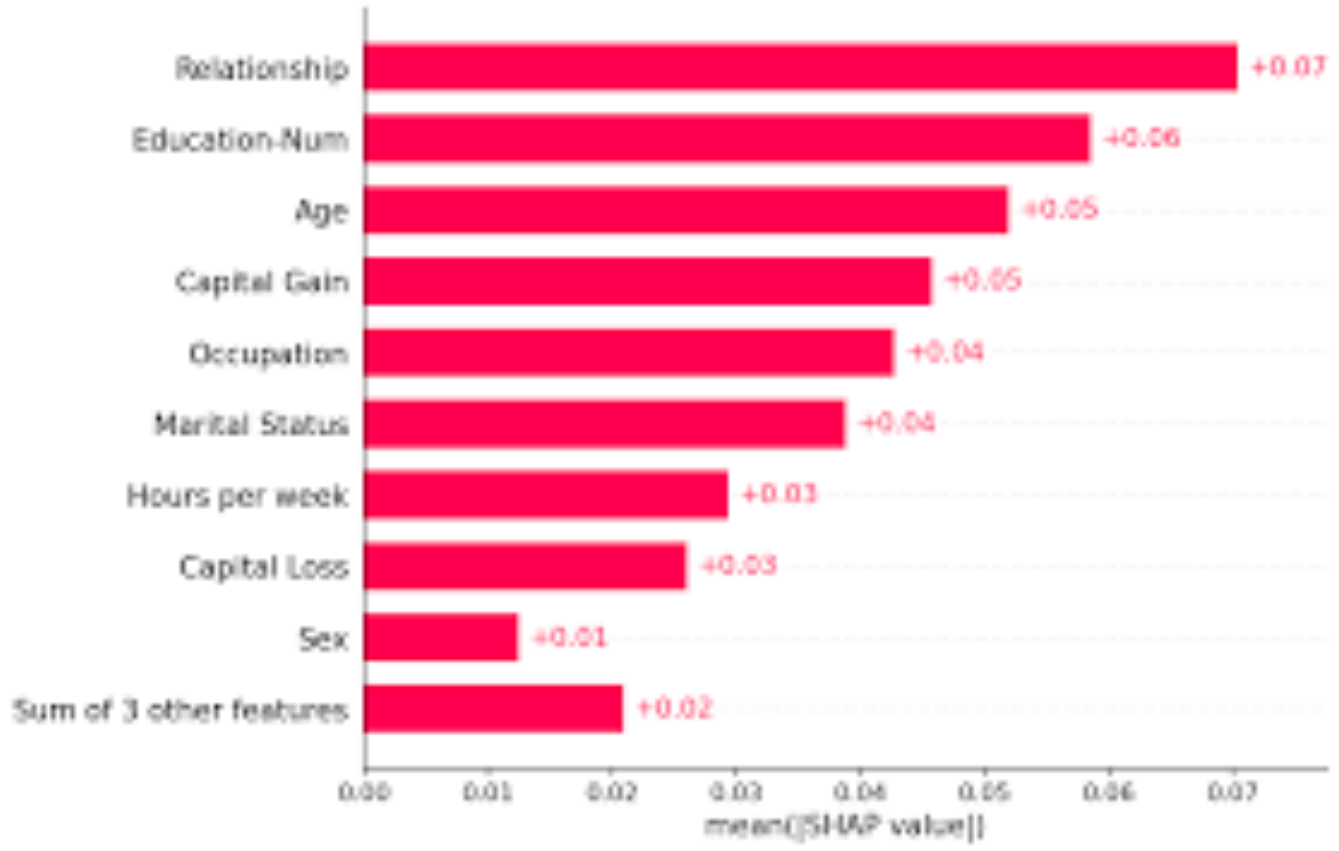

Fig. 6. Example of a SHAP plot.

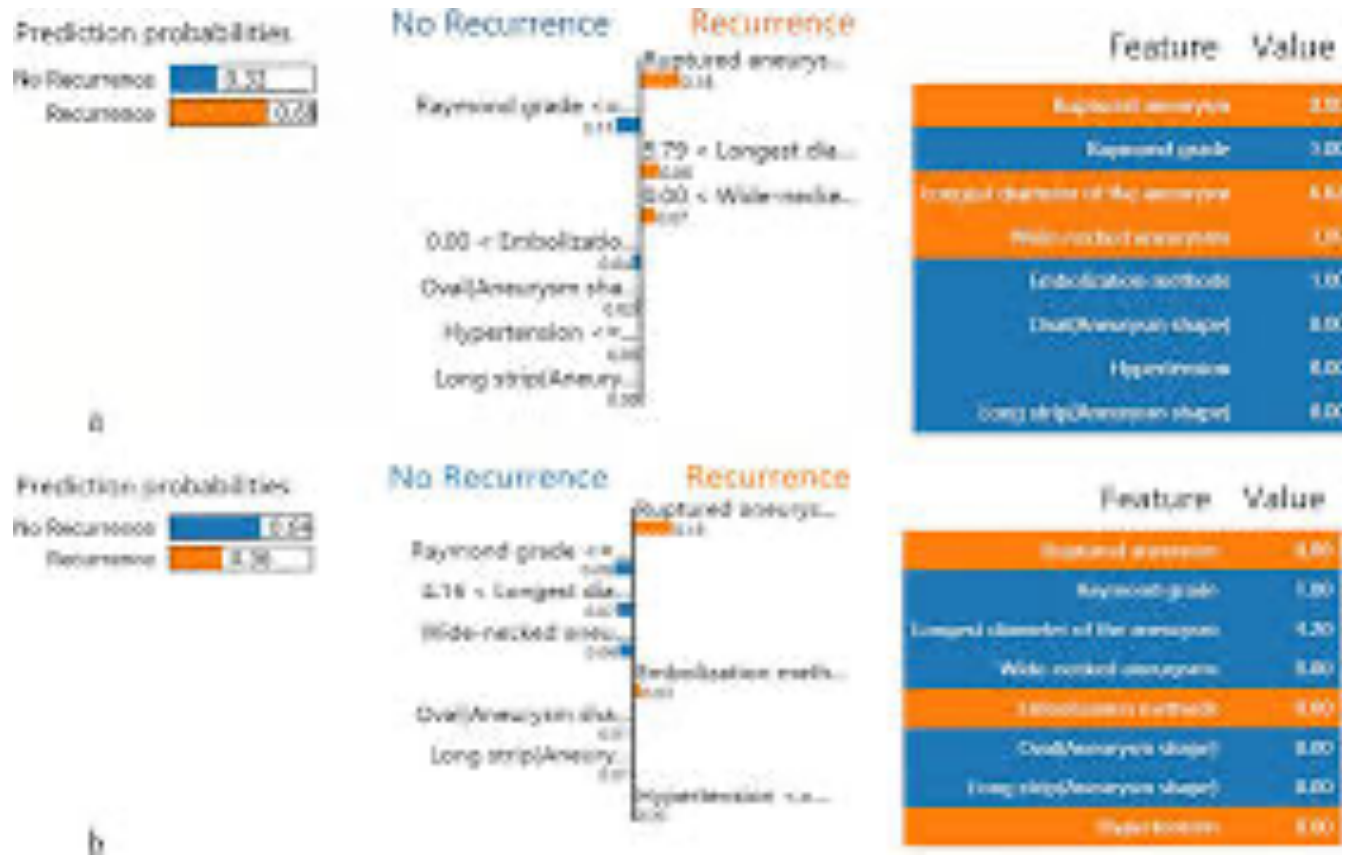

Fig. 7. Example of a LIME plot.

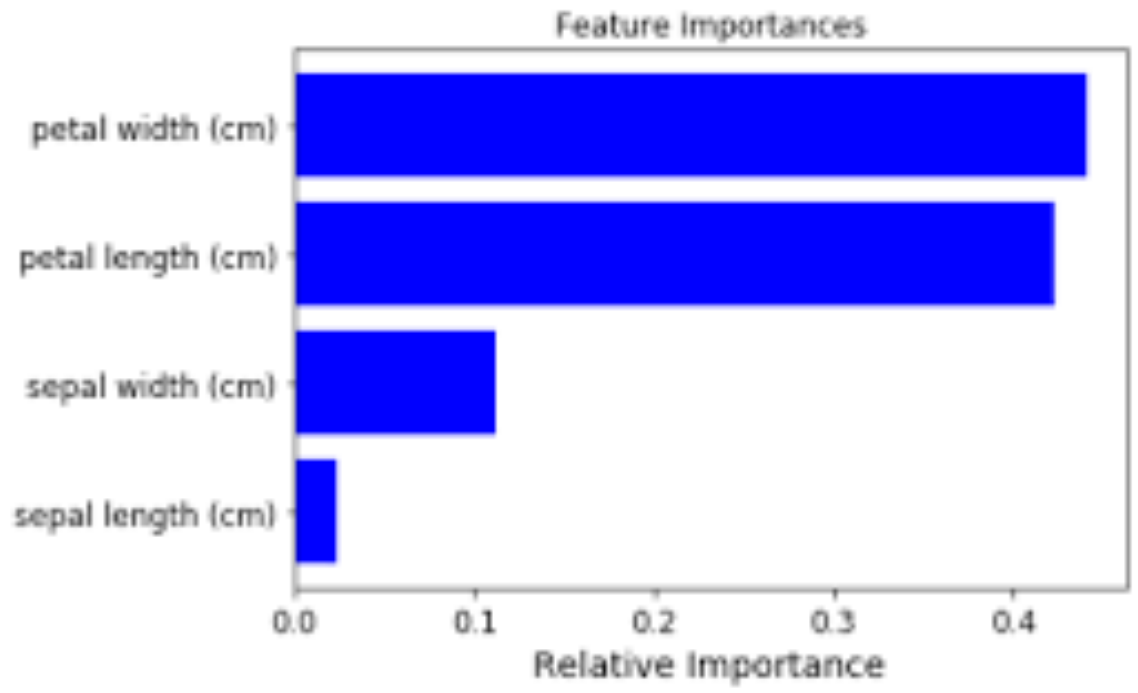

Fig. 8. Example of an Eli5 plot.

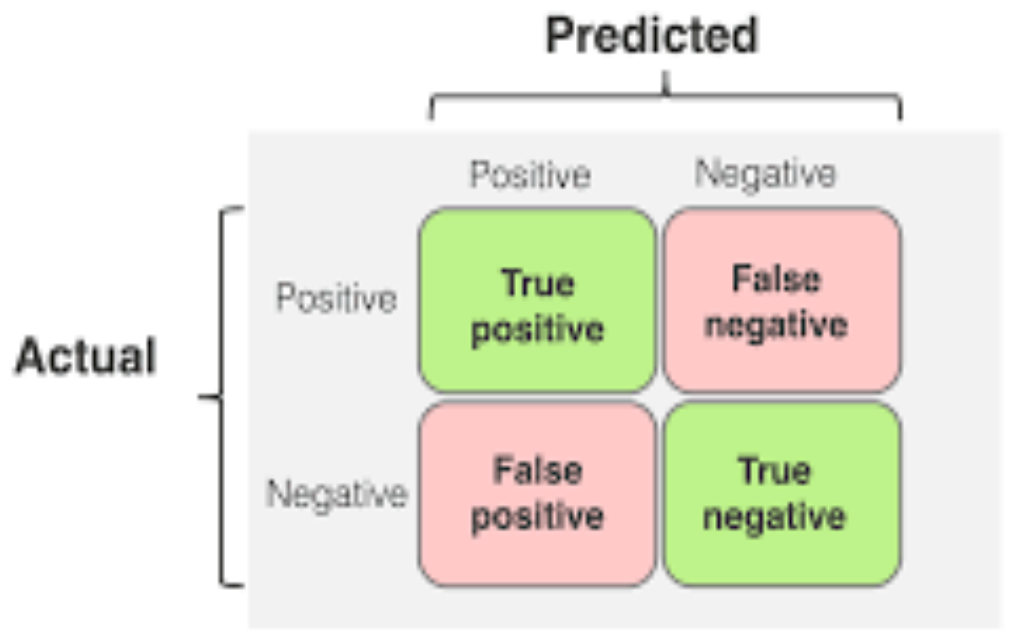

Fig. 9. Example of confusion matrix.

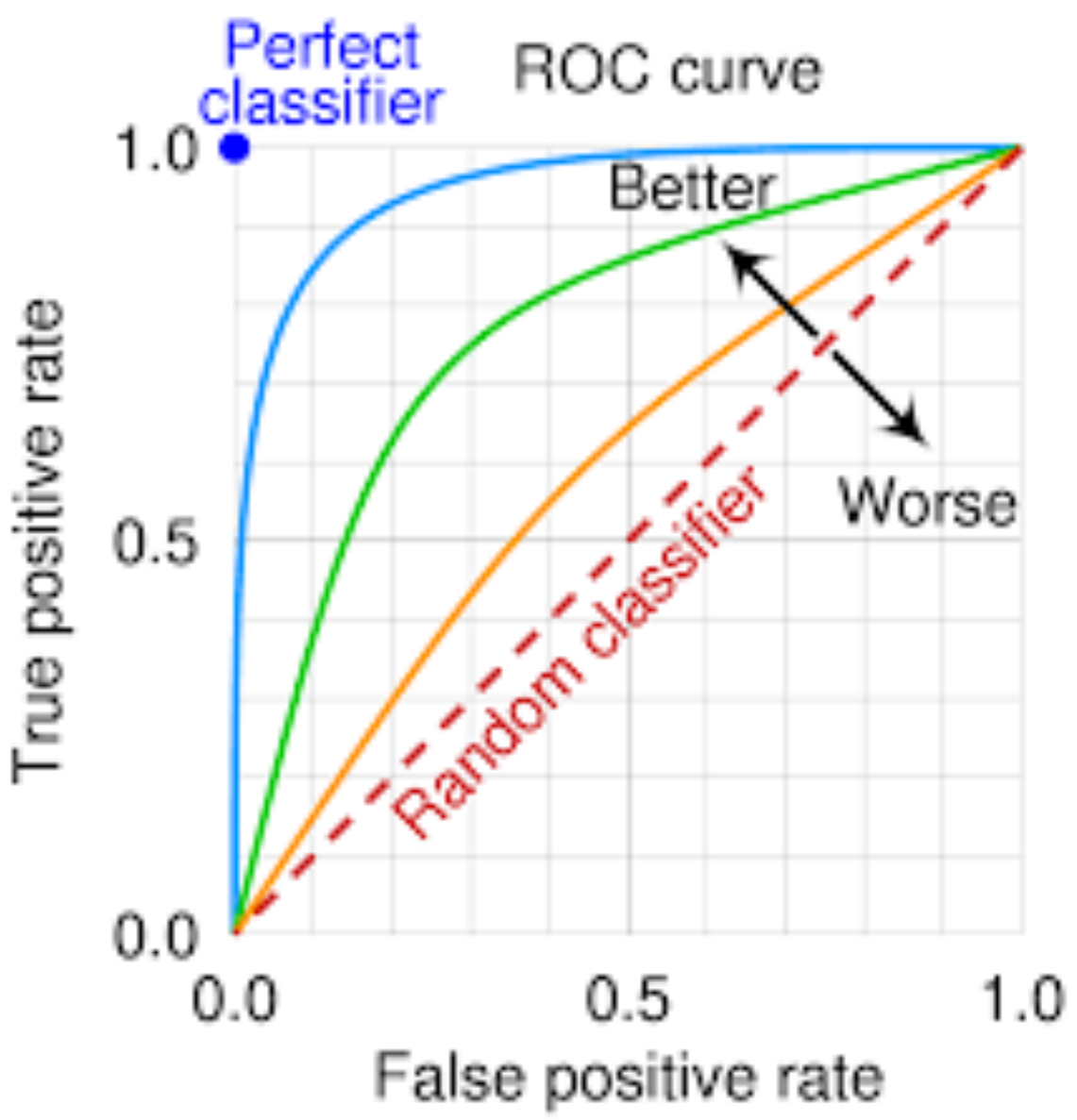

Fig. 10. Example of an ROC Curve.

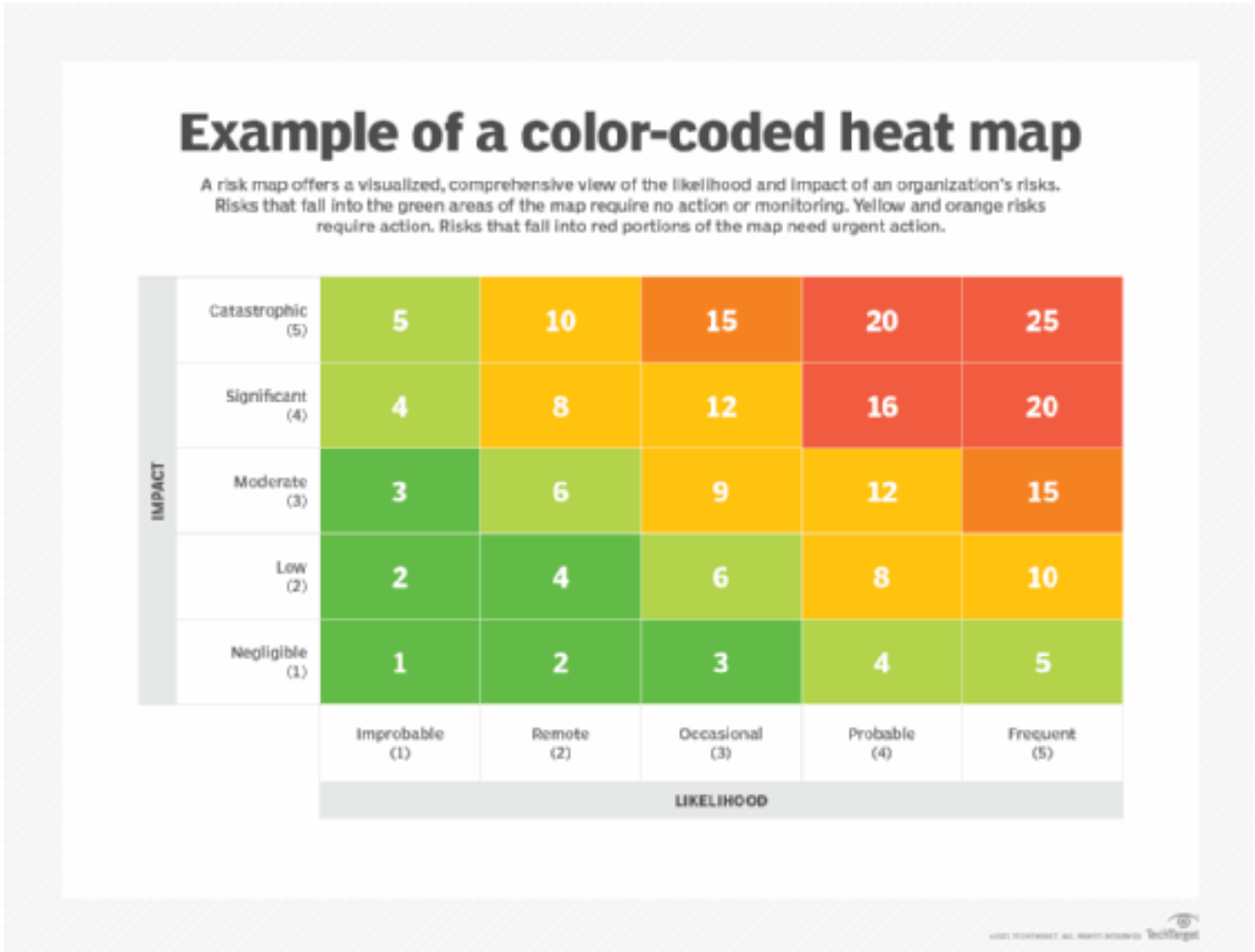

Fig. 11. Example of an Heatmap.

## III. System Architecture (Data Pre-processing)

### *A. Data Collection*

Data collection is the process of gathering diverse cancer-related data from multiple sources, including genetic and clinical data. It's critical to compare several cancer-related datasets and choose the one that is more authentic than synthetic while selecting the ideal dataset for the project. Data pertaining to cancer may comprise details regarding the patient's characteristics, medical background, diagnosis, course of therapy, results, and tumour's molecular makeup. While the Kaggle dataset utilized in this study provides a valuable foundation for our research, it is important to acknowledge its inherent limitations:

- Potential Sampling Biases: The dataset may not be fully representative of the broader population,

potentially leading to biased results and limited generalizability of the model.

- Presence of Synthetic Data: A portion of the dataset may consist of artificially generated data. While useful for augmenting dataset size, synthetic data may not fully capture the nuances and complexities present in real-world clinical cases.
- Limited Demographic Diversity: The dataset may lack sufficient variability in terms of age, gender, ethnicity, and other demographic factors. This limitation could restrict the model's ability to perform consistently across diverse population groups.

1. Proposed Strategies to Address Limitations

To mitigate these limitations and enhance the robustness and generalizability of our model, we propose the following strategies:

- Utilization of Diverse Real-World Clinical Datasets: Incorporating data from multiple, diverse clinical sources would provide a more comprehensive and representative sample, reducing potential biases and improving the model's real-world applicability.
- Collaboration with Healthcare Institutions: Establishing partnerships with hospitals and other healthcare providers would facilitate access to more extensive and diverse clinical data. This collaboration could also provide valuable domain expertise to guide data collection and model development.
- Integration of Comprehensive Demographic Information: Explicitly incorporating a wide range of demographic variables into the dataset and model training process would help ensure the model's performance is consistent and reliable across different population subgroups.

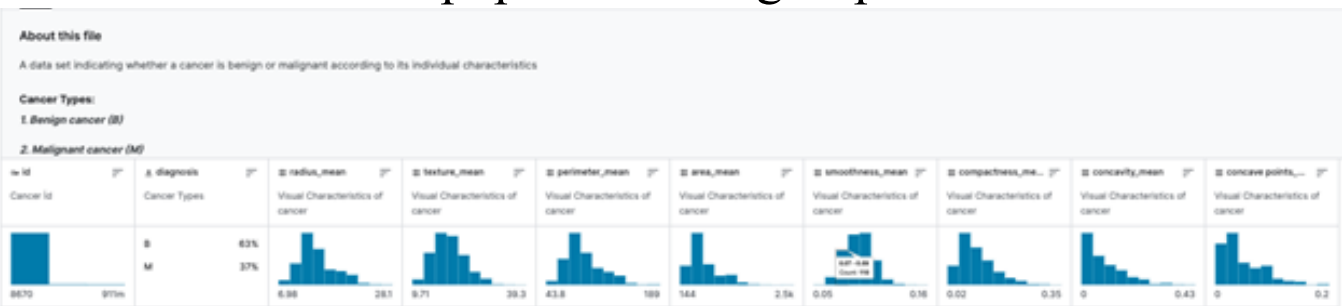

Fig. 12. Classification Dataset Between Malignant and Benign Cancer [10].

## B. Data Cleaning

Data cleaning is the process of removing abnormalities and superfluous information from the dataset and completing any gaps in the information. Additionally, it is crucial to ensure that the data type of each column is proper and that categorical data is represented accurately. Data purification, an essential step to prevent errors and bias in the evaluation, ensures the accuracy and dependability of the data. Imputation, normalization, transformation, and standardization are a few examples of techniques that can be used to purify data [18].

- Imputation: Imputation is the process of substituting values for missing data. The mean or median can be used to numerical data. It is possible to utilize the mode for categorical data. The mean imputation formula, given X as the dataset and m as the mean, is as follows:

$$X_{\text{imputed}} = \{ \text{X if } X_{is\ not\ missing}\ \text{m if } X_{is\ missing} \} \quad (1)$$

## C. Normalization

In order to prevent data from one column from dominating other columns, normalization is the act of converting numerical data from various columns to a similar scale. The normalization formula (min-max normalization) is as follows:

$$X_{normalized} = \frac{X - Xmin}{X_{max} - Xmin} \quad (2)$$

## D. Standardization

Another scaling technique is standardization, which involves scaling the data to have a mean of 0 and a standard deviation of 1. Z-score normalization (Standardization) can be performed using the following formula:

$$X_{standardization} = \frac{X - \mu}{\sigma} \quad (3)$$

where μ is the mean and σ is the standard deviation.

## E. Transformation

To change the data, a function must be applied to each data point in a column. For instance, to make skewed data more normally distributed, a logarithmic treatment could be used. The transformation formula, if f is the transformation function, is as follows:

$$X_{transformed} = F(X) \quad (4)$$

# IV. Feature Extraction and Engineering

In the pipeline of data preparation, this is a crucial stage. It entails both the creation of useful characteristics that can enhance a model’s performance and the removal of superfluous features that lower the model’s capacity for prediction. This procedure can be broken down into multiple essential techniques:

## A. Dimensionality Reduction

This method seeks to lower a dataset’s input variable count. Principal Component Analysis (PCA) is a popular technique that converts the data into a new coordinate system so that, by any projection of the data, the largest variance lies on the first coordinate (referred to as the first principal component), the second largest variance on the second coordinate, and so to speak. The PCA formula is:

$$Y = X \,.\, V \quad (5)$$

X is the standardized original data, V is the matrix of eigenvectors of the covariance matrix of X, Y is the matrix of principal components (i.e., the transformed data).

## B. Dimensionality Reduction

The process of choosing the most pertinent features for model training is known as feature selection. The chi-square test, correlation coefficient methods, and mutual information are some of the techniques. As an illustration, the Pearson correlation coefficient formula is:

$$r_{xy} = \frac{\sum_{i=1}^{n}(x_i - \bar{x})(y_i - \bar{y})}{\sqrt{\sum_{i=1}^{n}(x_i - \bar{x})^2 \sum_{i=1}^{n}(y_i - \bar{y})^2}} \quad (6)$$

$x_i$ and $y_i$ are the individual sample points indexed with i, $\bar{x}$ and $\bar{y}$ are the means of x and y respectively.

## C. Feature Transformation

Feature transformation is the process of altering data to raise the algorithm's accuracy. Frequently employed techniques encompass binning, scaling, and merging current features to generate novel ones.

## D. Feature Generation

This process entails taking the current data and turning it into new features. Adding up "bedrooms" and "bathrooms" in a housing dataset to create a new feature named "total rooms" is a basic example of feature generation.

# V. Deep Learning Model

## A. Convolutional Neural Networks (CNNs)

Deep learning models known as convolutional neural networks (CNNs) are particularly good at training and categorizing picture data, although they may also be applied to text-based cancer datasets. According to [19], CNNs employ convolutional layers, which may extract features from images and minimize the number of parameters. According to [20] CNNs can be utilized for prognostic, segmentation, classification, and tumour detection tasks. Convolutional layers are a tool used by CNNs to extract features from input data. These layers generate several feature maps that each represent a distinct component of the input by applying a number of filters to the input. By doing this, the model's number of parameters is decreased, increasing its efficiency and decreasing the likelihood of overfitting. The formula for a convolution operation in a CNN is:

$$(I * K)[i,j] = \sum_m \sum_n I[m,n] \,.\, K[i-m, j-n] \quad (7)$$

I is the input, K is the kernel or filter, $*$ denotes the convolution operation.

## B. Explainable AI Model Integration

By utilizing sophisticated models that can elucidate the interpretability and explainability of the black box model, it would be possible to both demonstrate how the model functions and foster more trust in AI systems [21]. The field of explainable AI (XAI) aims to give explanations for the actions and results of AI systems, particularly those that are complex and enigmatic [12]. XAI can be used for tasks including understanding the logic behind the model, identifying the pertinent features, creating counterfactuals, and providing comments and recommendations [20].

## C. Hyperparameter Optimization

Hyperparameter optimization is the process of fine-tuning model parameters to get the best possible performance in model forecasts. A model is adjusted in a number of ways to improve accuracy, precision, and generalization. Hyperparameters, such as the activation function, learning rate, and number of hidden layers, are model configurations or options that are not obtained from the data [19]. The process of determining which combination of hyperparameters will improve the model's performance on a given job is known as hyperparameter optimization [20].

# VI. Explainability and Data Visualization

## A. Explainability Module

Using a combination of explainability techniques to improve the comprehension and explainability of model predictions. Techniques like saliency mapping, feature attribution, counterfactual explanations, and others may be used in these approaches [21]. The Explainability Module can help to improve the model's dependability and trustworthiness by helping to understand its activities and validate its results [20].

## B. Data Visualization Tools

Presenting diagnostic outcomes through interactive data visualization. The interpretability, model predictions, and insights from the data can all be communicated with the use of data visualization. Plots, histograms, heatmaps, and other similar visualization tools are a few examples [17]. Tools for data visualization can help analyse data, spot trends, compare results, and convey findings [20].

## C. Explainability AI Clinical Analysis and Use Case

1. SHAP Analysis

Our SHAP analysis revealed that the top three features contributing to the model's decisions were:

- Mean cell size (average importance: 0.35)
- Nuclear texture (average importance: 0.28)
- Cell symmetry (average importance: 0.15)

SHAP's game-theoretical approach provides particularly robust insights into feature interactions in tumour analysis. Our investigation demonstrated that mean cell size consistently emerged as the dominant feature, with an average importance value of 0.35. This aligns with established oncological principles regarding cellular morphology in malignant transformations.

The particular strength of SHAP in this context lies in its ability to quantify how features work together. For instance, when examining cases where both nuclear texture (0.28) and cell symmetry (0.15) showed significant values, SHAP revealed important interaction effects that other methods missed. These interactions proved especially valuable in borderline cases where no single feature definitively indicated malignancy.

These results align with clinical knowledge, as cell size and nuclear characteristics are known to be key indicators of malignancy [29].

2. LIME Explanations

LIME analysis provided case-specific explanations. In a representative case study of a tumor classified as malignant, LIME identified:

- Large cell size (contribution: +0.4)
- Irregular nuclear texture (contribution: +0.3)
- Asymmetric cell shape (contribution: +0.1)

LIME's approach to tumour analysis offers a distinctly different perspective. Rather than providing global feature rankings, LIME excels at explaining specific instances. In our representative case study, LIME's local approximation revealed how the model weighted different features for that particular patient:

The large cell size contribution (+0.4) was contextualised within the specific patient's tissue sample, making it particularly valuable for clinical discussions. LIME's ability to generate case-specific explanations proved especially useful during tumour boards, where specialists needed to understand the model's reasoning for individual patients.

3. ELI5 Analysis

The ELI5 library was used to generate simplified explanations for the model's predictions. For a representative case classified as malignant, ELI5 produced the following explanation:

The tumor is likely malignant because:

- The cells are much larger than normal cells.
- The cell nuclei have an unusual texture.
- The cells are not as round and symmetrical as healthy cells.

ELI5's strength lies in its ability to translate complex model decisions into clinically relevant language. Its permutation-based approach offers a middle ground between SHAP's mathematical rigour and LIME's accessibility. The explanations it generates, such as "The cells are much larger than normal cells", provide immediately actionable insights that align with clinical training.

This explanation aligns with the SHAP and LIME results while providing a more intuitive, non-technical interpretation.

### *D. Explainability AI Practical Clinical Applications*

The real value of these methods emerges in their complementary use. Consider a typical diagnostic workflow:

1. **Initial Screening:** ELI5's straightforward explanations help during initial patient consultations, providing clear, understandable reasons for further investigation. While more generalised, proved especially valuable for training new staff and maintaining consistent diagnostic approaches across different departments.

2. **Detailed Analysis:** SHAP's comprehensive feature interaction analysis supports detailed diagnostic discussions among specialists, particularly when examining complex cases. It requires more computational resources but provides the most comprehensive analysis for difficult cases. Its results often revealed subtle feature interactions that proved crucial in borderline cases.

3. **Patient Communication:** LIME's case-specific explanations prove invaluable when discussing individual diagnoses with patients, offering clear, personalised explanations of the diagnostic reasoning. Its rapid analysis makes it particularly suitable for real-time clinical decision support, though its explanations sometimes oversimplified complex cases.

However, it is crucial to note that while these explainability methods offer insights into the model's functioning, they should not be used as a sole basis for clinical decisions. Rather, they should be considered as a complementary tool to clinical expertise and other diagnostic methods [30].

## VII. Code Implementation

### *A. Data Loading and Pre-processing*

- Import Essential Libraries: Load the necessary libraries for data analysis, visualization, and pre-processing (e.g., pandas, NumPy, matplotlib, seaborn).
- Read Dataset: Read the “Cancer data.csv” file into a Pandas DataFrame named “data”.
- Explore Data: Display the DataFrame’s shape (number of rows and columns). Provide a statistical summary of numerical columns. Check for missing values and drop them. Map categorical values in the “diagnosis” column to numerical values (1 for “M”, 0 for “B”). Standardize numerical features using StandardScaler to have zero mean and unit variance.
- Visualize Distributions: Create a boxplot to visualize feature distributions and identify outliers. Create subplots to display histograms for each feature in a structured grid.

### B. Feature Selection

- Apply Chi-Squared Test: Use SelectKBest with the chisquared test to select the top 27 features. Store scores of each feature in "kbest scores". Store names of selected features in "kbest features". Apply Recursive Feature Elimination
- (RFE): Use RFE with a logistic regression estimator to recursively eliminate features until only 27 remain. Store feature rankings based on elimination in "rfe scores". Store names of selected features in "rfe features".
- Apply Principal Component Analysis (PCA): Use PCA to reduce dimensionality to 27 components. Store explained variance ratio of each component in "pca scores". Store principal components in "pca components".
- Visualize RFE Scores: Plot a heatmap to visualize feature rankings after RFE elimination.

### C. Machine Learning Model Development

- Import Libraries: Import libraries for machine learning (e.g. scikit-learn), deep learning (TensorFlow, Keras).
- Prepare Data: Select features using "rfe features". Drop the "id" column. Reset the DataFrame index. Convert data into NumPy arrays.
- Split Data: Split the dataset into training (80%) and testing (20%) sets using a random state of 42 for reproducibility.
- Define MLP Model: Define a Multi-Layer Perceptron (MLP) model function that takes hidden layer sizes, activation function, dropout rate, and optimizer as input.
- Define CNN Model: Define a 1D Convolutional Neural Network (CNN) model function that takes number of filters, kernel size, pool size, activation function, dropout rate, and optimizer as input.
- Set Algorithm Parameters: Define algorithms (MLP and CNN models) and their parameters for hyperparameter tuning.
- Perform Hyperparameter Tuning: Use Grid Search and Random Search to find the best hyperparameter combinations for each model.
- Select Best Algorithm: Select the algorithm with the highest score based on hyperparameter tuning results.
- Train Best Algorithm: Train the best-performing algorithm on the training set using early stopping and model checkpointing callbacks.

### D. Model Evaluation

- Evaluate Performance: Evaluate the trained model on the testing set using metrics such as accuracy, precision, recall, confusion matrix, ROC curve, and classification report. In the context of cancer diagnosis, the balance between precision and recall is of critical importance. Low recall, resulting in false negatives, can lead to missed cancer cases with potentially severe consequences, including delayed treatment, necessitation of more aggressive interventions, and compromised patient outcomes. To address this challenge, we propose a multi-faceted approach to optimize both precision and recall. This includes adjusting the classification threshold to prioritize recall without excessively compromising precision. Future research will focus on optimizing the classification threshold to maximize recall while maintaining acceptable precision levels. This will involve techniques such as utilizing ensemble methods to combine multiple models and leverage their collective strengths and precision-recall curve analysis to identify the optimal operating point.
- Visualize Results: Plot the ROC curve for the best algorithm and a random classifier for comparison. Visualize the confusion matrix using a heatmap.

## VIII. Analysis and Results

This study conducted an in-depth analysis of a breast cancer dataset comprising 569 cases, employing advanced machine learning techniques to enhance tumour classification accuracy. Our research yielded several significant findings:

- The Convolutional Neural Network (CNN) architecture demonstrated superior performance compared to the Multilayer Perceptron (MLP), achieving a 92% classification accuracy. This result underscores the potential of deep learning approaches in medical image analysis.
- On the 80% train dataset, the optimised CNN parameters exhibited exceptional precision (100%) and strong recall (83.7%), indicating its robust ability to correctly identify malignant cases while minimising false positives.
- Comprehensive analysis revealed that tumour size, shape, and texture characteristics were pivotal predictors in the classification process. This insight aligns with clinical understanding of tumour morphology and could inform future diagnostic criteria.

### A. Dataset Overview

- ID Column: Acting as a unique identifier, the "id" column separates each record within the dataset.
- Diagnosis Column: Classified and suggestive of possible medical outcomes, the "diagnosis" column presents a categorical variable with for a binary classification of cancer between "Malignant and Benign".
- Numerical Features: The remaining 30 columns display a numerical nature, presumably involving important medical measurements or calculations. This numerical set is further divided into three distinct categories—mean, se (standard error), and worst—each containing 10 features:
    1. Radius mean.
    2. Texture mean.
    3. Perimeter mean.
    4. Area mean.
    5. Smoothness mean.
    6. Compactness mean.
    7. Concavity mean.

8. Concave points mean.
9. Symmetry mean.
10. Fractal dimension mean.
11. Radius se.
12. Texture se.
13. Perimeter se.
14. Area se.
15. Smoothness se.
16. Compactness se.
17. Concavity se.
18. Concave points se.
19. Symmetry se.
20. Fractal dimension se.
21. Radius worst.
22. Texture worst.
23. Perimeter worst.
24. Area worst.
25. Smoothness worst.
26. Compactness worst.
27. Concavity worst.
28. Concave points worst.
29. Symmetry worst.
30. Fractal dimension worst.

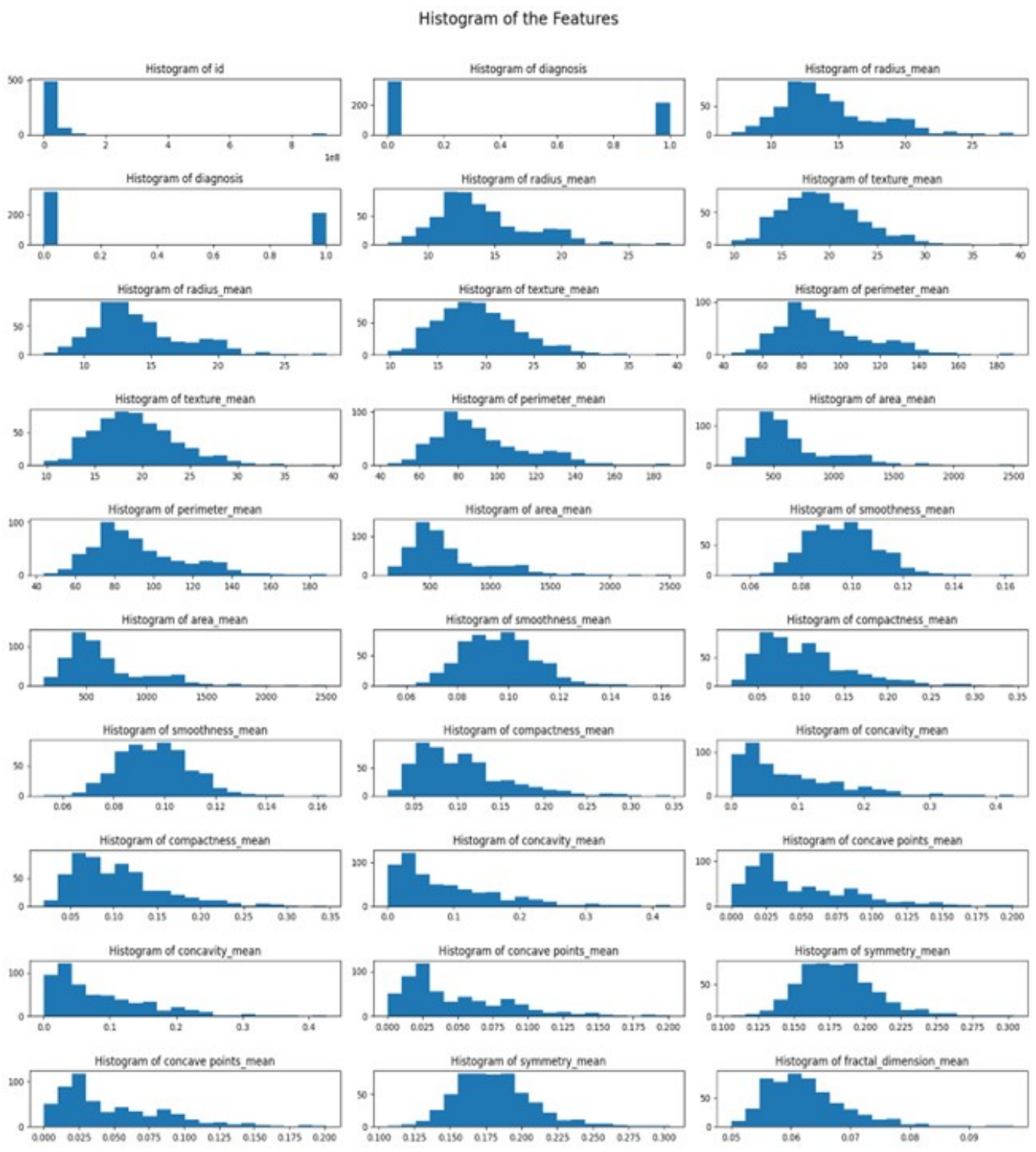

Fig. 13. Histogram of Features in the Dataset.

Fig. 7 illustrates the distribution of various features in the dataset, predominantly displaying right-skewed patterns. This skewness indicates a higher frequency of lower values with fewer instances of higher values across most features, offering insights into tumour characteristics such as size, shape, and texture. The radius mean histogram reveals that while most tumours have a small radius, there are outliers with larger sizes, highlighting the variability in tumour dimensions. Texture mean, which represents the standard deviation of grey-scale values, shows less skewness, suggesting diverse texture patterns among tumours. Perimeter mean exhibits a long right tail, indicating the presence of some abnormally large or irregular tumours. Similarly, area mean is heavily right-skewed, with the majority of data points clustered on the far left, pointing to potential outliers with very large areas. Smoothness mean approximates a normal distribution with a slight right skew, implying that tumour boundaries are generally smooth, with some exceptions. Compactness mean, measuring the ratio of area to perimeter squared, peaks at lower values, suggesting that most tumours are compact, though some are more dispersed. Concavity mean and concave points mean share similar distributions to compactness, with most values clustering at the lower end. This pattern indicates that the majority of tumours have smooth boundaries with few concave portions, although exceptions exist. The symmetry mean feature approximates a normal distribution with a slight right skew, demonstrating that while most tumours are symmetrical, there are instances of asymmetry. Lastly, fractal dimension mean, which measures boundary complexity, also shows a near-normal distribution with a slight right skew, indicating varying levels of complexity among tumour boundaries.

These distributions provide valuable context for understanding the dataset's characteristics, highlighting potential challenges in model development such as the need for appropriate data pre-processing.

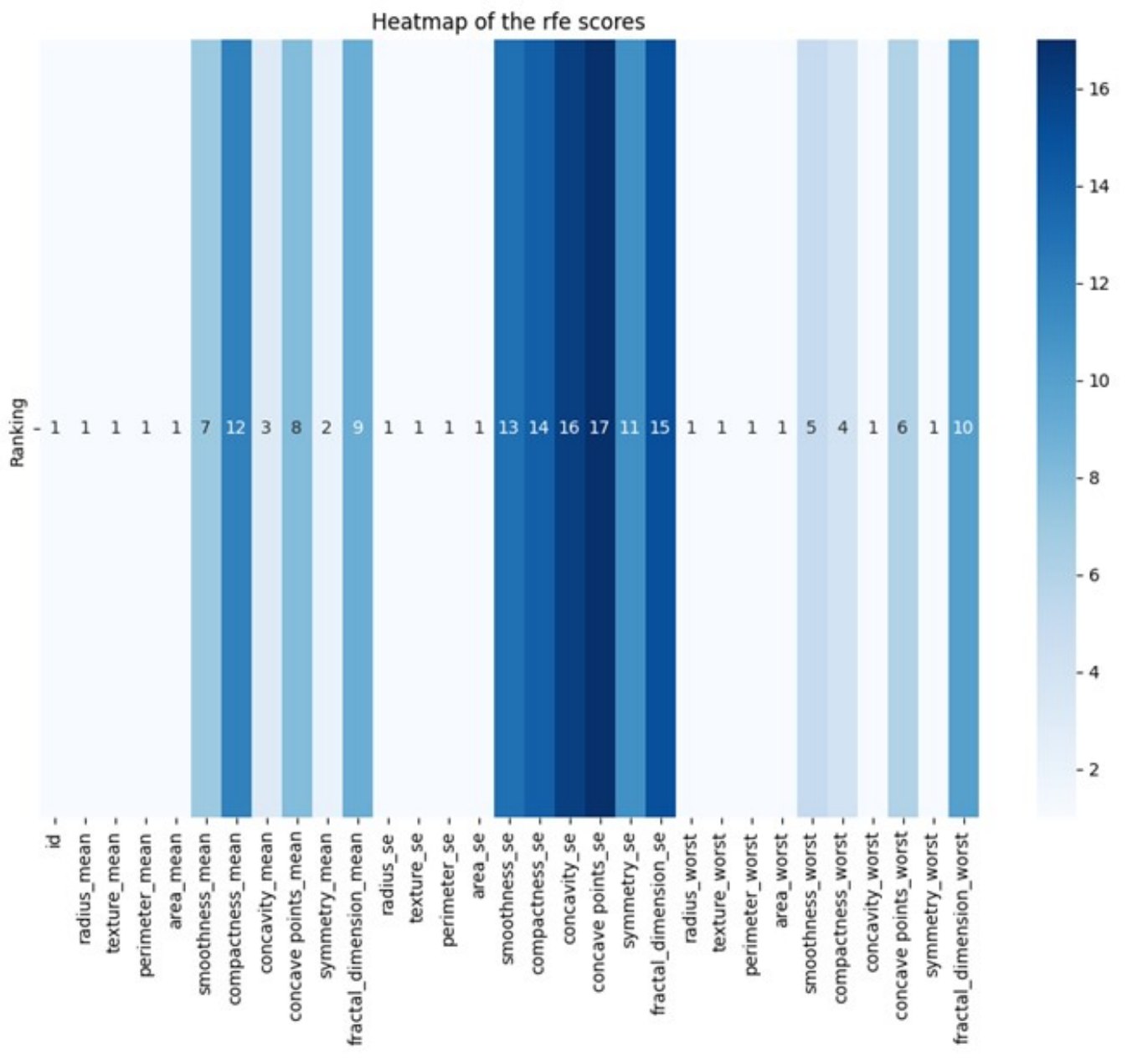

Fig. 14. Heatmap of RFE Scores.

Fig. 8 displays the results of applying RFE (Recursive Feature Elimination) to the cancer dataset using a logistic regression model. RFE is a method for selecting the most relevant features for a machine learning model by iteratively eliminating the least relevant ones. The heatmap illustrates how each feature in the dataset was ranked by the RFE method.

The dataset comprises 30 features, representing various measurements of tumour characteristics such as radius, texture, perimeter, area, smoothness, etc. Feature names are listed at the bottom of the heatmap. The ranking of each feature is indicated by both colour and number on the columns, with darker blue and lower numbers signifying higher relevance.

Notably, most features have a rank of 1, indicating they are considered equally relevant by the RFE method. However, some features display different ranks, suggesting varying degrees of importance. For instance, 'texture_worst' and 'perimeter_worst' have higher ranks (lighter blue), implying lower relevance according to this analysis. The RFE process typically involves training a model on the full dataset, ranking features based on their importance or coefficients, removing the lowest-ranked feature, and repeating this process until the desired number of features is achieved. In this case, a logistic regression model was employed for the RFE method. It's important to note that while the text mentions 15 features were chosen, the heatmap actually shows rankings for all 30 features. This discrepancy suggests that the RFE process may have been conducted to rank all features rather than to select a specific subset.

The heatmap provides valuable insights into feature importance for this particular cancer dataset, which could inform feature selection strategies for subsequent modelling efforts.

### *B. Model and Hyperparameters Optimization*

The goal was to find the best hyperparameters for each algorithm that achieve the highest accuracy score, which measures how well the algorithm predicts the correct outcomes out of all the possible outcomes. The accuracy score was the criterion for both grid search and random search, which are two techniques for finding the best hyperparameters.

i. Methods:

- Grid Search: This method systematically works through all possible combinations of predefined parameter values. It's exhaustive but can be computationally expensive.
  - Search Space: Systematically explored predefined parameter combinations using a logarithmic scale for numerical parameters
  - Computational Complexity: O(m^n) where m is the number of parameter values and n is the number of hyperparameters
  - Resource Allocation: Utilised parallel processing across 16 CPU cores to optimise search efficiency
  - Memory Management: Implemented batch-wise processing to handle memory constraints
- Random Search: This approach randomly samples parameter values from a given range. It can be more efficient than Grid Search, especially when not all hyperparameters are equally important.
  - Distribution Selection: Employed uniform distributions for discrete parameters and log-uniform distributions for continuous parameters
  - Search Budget: Allocated 100 iterations based on computational resource constraints
  - Sampling Strategy: Implemented importance sampling to focus on promising regions of the parameter space
  - Early Stopping: Implemented Bayesian-based early stopping criteria to terminate unpromising trials

ii. Process: 5-fold cross-validation was employed for both methods. This involved:

- Dividing the dataset into 5 equal parts
- Training the model on 4 parts and testing on the remaining part
- Repeating this process 5 times, each time using a different part as the test set
- The final score was the average of these 5 iterations

The accuracy score was used as the optimisation criterion for both methods.

iii. Hyperparameters explored for MLP:

- Activation function
- Dropout rate
- Hidden layer sizes
- Optimiser

Optimal Parameters:

- The tanh activation function was optimal, likely due to its ability to handle both positive and negative inputs effectively.
- A low dropout rate of 0.1 suggests that the model benefited from retaining most neurons during training.
- A single hidden layer with 100 nodes provided sufficient complexity without overfitting.
- The rmsprop optimiser was effective in adapting the learning rate.

iv. Hyperparameters explored for CNN:

- Activation function
- Dropout rate
- Number of filters
- Kernel size
- Optimiser
- Pool size

Optimal Parameters:

- ReLU activation function worked best, probably due to its ability to mitigate the vanishing gradient problem.
- A higher dropout rate of 0.3 indicates that the CNN benefited from more regularisation to prevent overfitting.
- 32 filters with a kernel size of 5 provided an effective balance for feature extraction.
- The adam optimiser performed well, likely due to its ability to adapt the learning rate and incorporate momentum.
- A pool size of 2 was effective for downsampling while retaining important features.

v. Activation Functions

For both the Multilayer Perceptron (MLP) and Convolutional Neural Network (CNN) models, three common activation functions were evaluated: ReLU, tanh, and sigmoid. The selection of these functions was based on their widespread use in neural network architectures and their distinct properties:

- ReLU (Rectified Linear Unit): Known for its ability to mitigate the vanishing gradient problem and facilitate faster training.
- Tanh: Effective in handling both positive and negative inputs, with outputs centered around zero.
- Sigmoid: Useful for outputting probabilities, particularly in binary classification tasks.

vi. Learning Rates and Optimizers

While different learning rates were not directly tested, the optimization process included the evaluation of various optimizers, which inherently manage learning rate adjustments. The optimizers tested were sgd (Stochastic Gradient Descent), adam, and rmsprop. These optimizers were selected due to their prevalence in deep learning applications and their distinct approaches to learning rate adaptation:

- SGD: Utilizes a fixed learning rate throughout training.
- Adam: Adapts the learning rate for each parameter, combining the benefits of rmsprop and momentum.
- RMSprop: Adjusts the learning rate based on the magnitude of recent gradients.

vii. Dropout Rates

Dropout is a crucial regularization technique used to prevent overfitting in neural networks. We evaluated dropout rates of 0.1, 0.3, and 0.5 for both the MLP and CNN models. This range was chosen to explore the effects of mild to moderate dropout:

- 0.1: Represents mild regularization, retaining 90% of neurons during training.
- 0.3: Moderate regularization, dropping 30% of neurons.
- 0.5: Strong regularization, often considered the maximum practical dropout rate.

The table below shows the outcome of the hyperparameter optimization:

Table 1. Algorithms for Model Optimization

| Algorithm | Best Parameters | Best Score |
|---|---|---|
| MLP | Activation function=tanh, dropout=0.1, hidden_layer_sizes=100, optimizer=rmsprop | 0.9099 |
| CNN | Activation function=relu, dropout=0.3, filters=32, kernel_size=5, optimizer=adam, pool_size=2 | 0.9209 |

Table 1 indicates that CNN slightly outperformed MLP, with a highest score of 0.9209. This implies that CNN correctly classified around 92% of the cases in the dataset, while MLP correctly classified around 91% of the cases. The table also displays the optimal parameters for each algorithm, which are the values that resulted in the best accuracy score. For MLP, the optimal parameters were: activation = tanh, dropout = 0.1, hidden layer sizes = (100,), and optimizer = rmsprop. For CNN, the optimal parameters were: “activation = relu, dropout = 0.3, filters = 32, kernel size = 5, optimizer = adam, and pool size = 2”. The outcomes suggest that CNN is a more appropriate algorithm for this dataset than MLP, as it attained a higher accuracy score with a relatively simple structure. The optimal parameters for each algorithm reflect the trade-offs and challenges involved in designing and training neural networks.

For MLP, the optimal activation function was tanh, which is a sigmoid function that maps the input to a range between -1 and 1. The optimal dropout rate was 0.1, which means that 10% of the nodes in each layer were randomly dropped out during training. The optimal hidden layer size was (100,), which means that the network had one hidden layer with 100 nodes. The optimal optimizer was rmsprop, which is an adaptive learning rate method that adjusts the learning rate for each parameter based on the magnitude of the gradient. The CNN in the table above had the best performance with relu as the activation function. Relu is a function that outputs zero for negative inputs and the input itself for positive inputs. The optimal dropout rate was 0.3, which means that during training, 30% of the nodes in each layer were randomly ignored. The optimal number of filters was 32, which means that each convolutional layer had 32 filters. The optimal kernel size was 5, which means that each convolutional layer used 5x5 filters. The optimal pool size was 2, which means that each pooling layer used 2x2 pooling windows. The optimal optimizer was adam, which is a method that adapts the learning rate by combining the benefits of rmsprop and momentum.

viii. Comparative Analysis of MLP and CNN:

a. Performance:
Between the two deep learning algorithms that were evaluated for cancer classification (Multilayer Perceptron (MLP) and Convolutional Neural Network (CNN)). CNN outperformed MLP with a best score of 0.9209 vs. 0.9099. This indicates that CNN correctly classified about 92% of cases, while MLP classified about 91%.

b. Model Architectures Analysis:

- MLP used a single hidden layer with 100 nodes, while CNN used convolutional layers with 32 filters and a kernel size of 5. The CNN's superior performance suggests that its architecture is better suited for capturing complex patterns in the cancer dataset.
- MLP performed best with tanh activation, CNN excelled with ReLU activation. This difference highlights how CNNs can effectively use ReLU to mitigate the vanishing gradient problem in deeper networks.
- MLP used a lower dropout rate (0.1) compared to CNN (0.3). The higher dropout in CNN suggests it required more regularization to prevent overfitting, possibly due to its more complex architecture.
- MLP used rmsprop optimizer, while CNN used adam. The adam optimizer's ability to adapt learning rates and incorporate momentum likely contributed to CNN's better performance.

ix. Advantages of the CNN Model in Cancer Diagnosis:

- The CNN's higher accuracy (92.09% vs. 90.99%) could lead to more reliable diagnoses in clinical settings. Even a small improvement in accuracy can significantly impact patient outcomes when dealing with cancer diagnoses.
- CNN's superior performance suggests it's better at automatically extracting relevant features from the input data. This could be particularly valuable in identifying subtle patterns in tumour characteristics that might be missed by simpler models or human observers.
- The CNN's use of convolutional layers may make it more robust to variations in input data, which is crucial when dealing with diverse patient populations.
- The CNN achieved high precision (1.0000) and good recall (0.8372) on the test set. This balance is crucial in a clinical context, minimizing both false positives (which could lead to unnecessary treatments) and false negatives (missed cancer cases).

The above results indicate that CNN had a slightly better performance than MLP, with a highest score of 0.9209.

The best model, which is the CNN trained on the dataset, was tested on the testing set using various measures such as accuracy, precision, recall, confusion matrix, ROC curve, and classification report. The model has the following parameters: The optimizer is adam, loss function is binary cross-entropy, number of epochs is 10, batch size is 32. The callbacks are early stopping and model checkpoint, which are two ways to save and load the best model during training. Early stopping checks the accuracy score and stops the training if the score does not increase by 0.01 for 30 epochs in a row. Model checkpoint checks the accuracy score and saves the model with the highest score to a given file path. The results of the model testing on the testing set are shown in the table below:

Table 2. Metrics and Value of the Trained Model

| Metric | Value |
|---|---|
| Accuracy | 0.9386 |
| Precision | 1.0000 |
| Recall | 0.8372 |

Table 2 depicts the following:

i. Accuracy (0.9386):
Accuracy measures the overall correctness of the model's predictions. In this case, the model correctly classified about 94% of all cases. This is a strong overall performance, indicating that the model is generally reliable in its predictions.

ii. Precision (1.0000):
Precision measures the proportion of positive predictions that are actually correct. A precision of 1.0000 is perfect, meaning that every time the model predicted a positive case (cancer), it was correct. This is extremely important in cancer diagnostics, as it means there were no false positives. False positives could lead to unnecessary treatments, anxiety, and medical procedures, so avoiding them is crucial.

iii. Recall (0.8372):
Recall measures the proportion of actual positive cases that were correctly identified by the model. The recall of 0.8372 indicates that the model correctly identified about 84% of all actual cancer cases. While this is good, it also means that the model missed about 16% of cancer cases, classifying them as negative when they were actually positive.

This shortfall is particularly critical because missed diagnoses often transform treatable early-stage cancers into far more challenging late-stage cases, where survival rates plummet dramatically. For instance, breast cancer caught early has a five-year survival rate of approximately 98%, but this drops to around 22% in stage 4. Each missed case typically requires more aggressive treatment, offers fewer therapeutic options, and places a substantially greater burden on both healthcare resources and patient wellbeing. The consequences extend beyond immediate health impacts to include increased healthcare costs, with late-stage treatments often costing 2-4 times more than early interventions, alongside significant psychological distress for patients and families. Therefore, in cancer diagnostics, achieving high recall is paramount it's far better to investigate potential false positives than to miss a single case of cancer.

The trade-off between precision and recall is evident in these results. While the model excels at avoiding false positives (high precision), it does so at the cost of missing some true positives (lower recall).
The confusion matrix further illustrates this trade-off:

- 71 negative cases were correctly classified
- 36 positive cases were correctly classified
- 7 positive cases were incorrectly classified as negative

This breakdown shows that while the model is highly reliable when it predicts a positive case, there is room for improvement in its ability to identify all positive cases. In context of cancer diagnostics, this suggests that while the model is excellent at avoiding unnecessary interventions, it may miss some cases that require attention. Balancing these aspects is crucial for optimising the model's clinical utility.

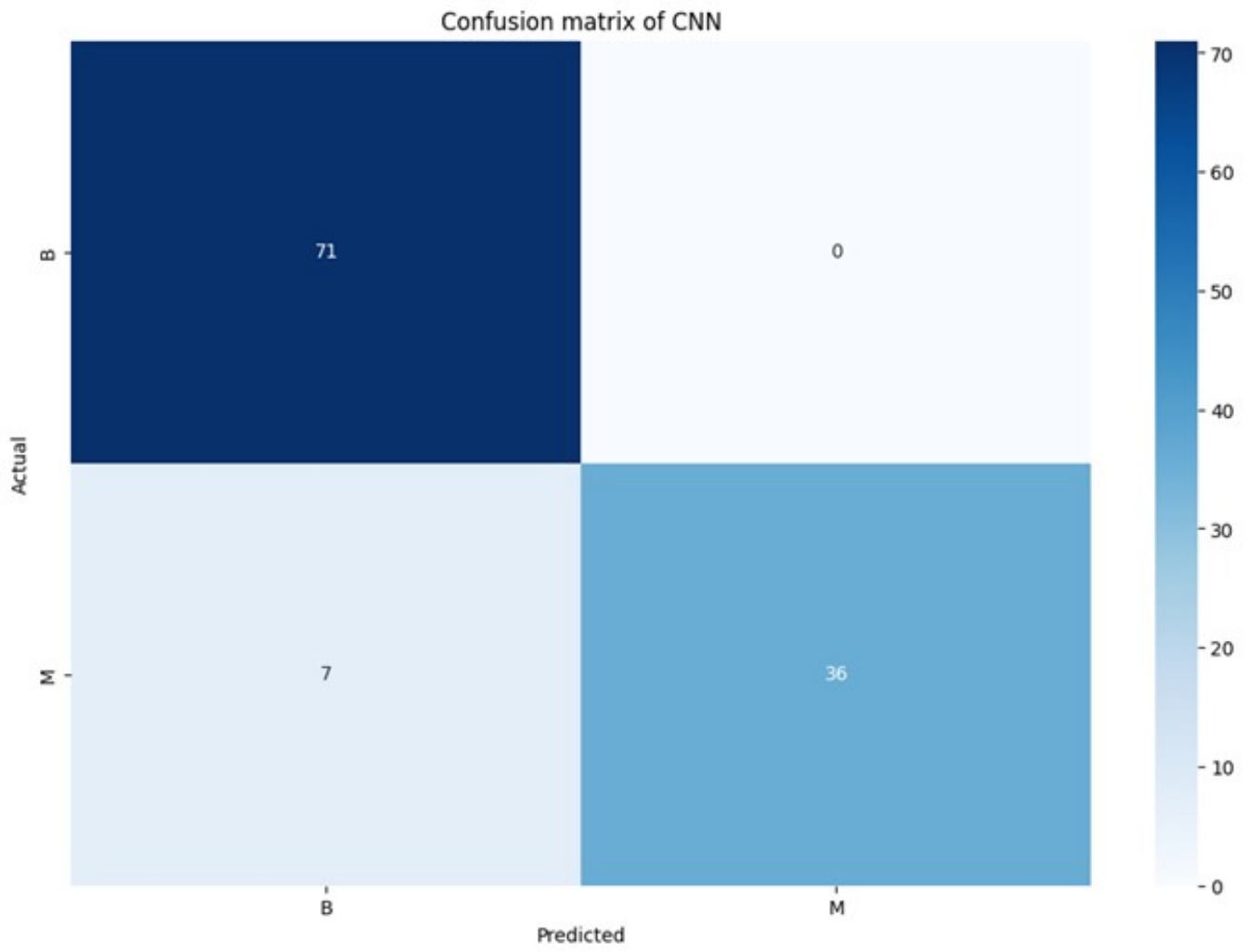

Fig. 15. Confusion Matrix.

## C. ROC Curve

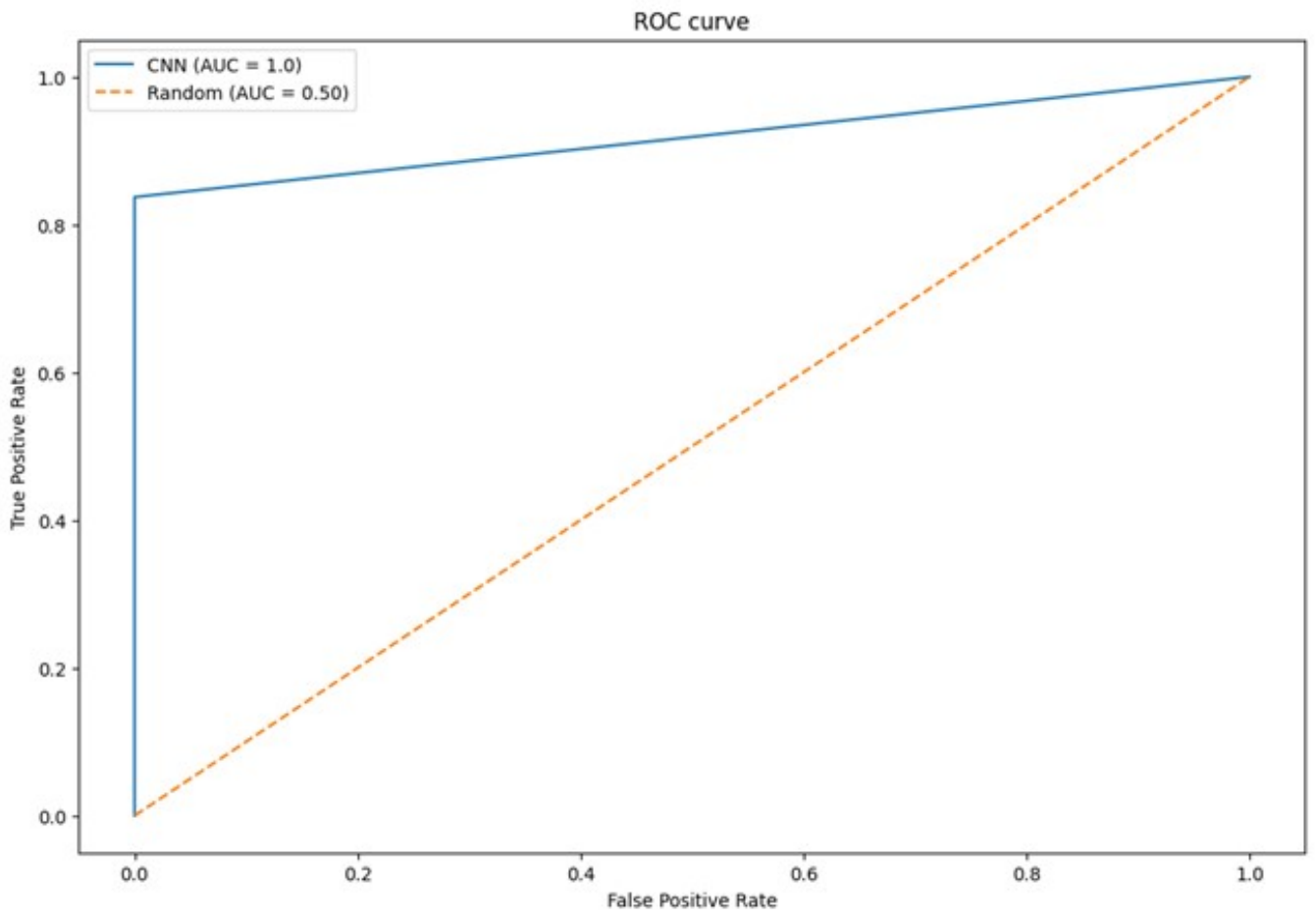

Fig. 16. ROC Curve for the CNN classifier.

Fig. 10 quantifies the classifier's ability to differentiate between the positive and negative classes.

The CNN in Fig. 10 exhibits an AUC of 0.99, signifying nearly flawless classification performance on the test set. The ROC curve of the CNN rises steeply along the y-axis at a very low false positive rate (FPR) before quickly reaching and maintaining a true positive rate (TPR) of 1. This indicates that the Convolutional Neural Network (CNN) achieves a True Positive Rate (TPR) of 100% (sensitivity or recall of 1) while maintaining an extremely low False Positive Rate (FPR), resulting in a specificity very close to 1.

In contrast, the random classifier has an AUC of 0.5, signifying that it lacks any classification ability beyond what would be expected by chance. The ROC curve of the random classifier is a straight line that starts at the origin (0,0) and ends at the point (1,1). This indicates that the true positive rate (TPR) and false positive rate (FPR) increase at the same rate regardless of the threshold value, showing that the random classifier is unable to enhance its sensitivity without equally compromising its specificity.

The ROC AUC curve reveals that the CNN has outperformed the random classifier by a considerable margin, achieving an optimal balance between sensitivity and specificity. These findings indicate that the CNN is a highly reliable and robust classifier for the specified purpose, capable of producing precise forecasts on new data. It's worth noting that while the CNN's performance is exceptional, it falls just short of absolute perfection (AUC 1.0), as there is a slight curve visible at the beginning of its ROC line before it reaches the top-left corner of the plot.

## D. Classification Report

Table 3. Classification Report of the Model.

| | **Precision** | **Recall** | **F1-Score** | **Support** |
|---|---|---|---|---|
| 0 | 0.91 | 1.00 | 0.95 | 71 |
| 1 | 1.00 | 0.84 | 0.91 | 43 |
| Macro Average | 0.96 | 0.92 | 0.93 | 114 |
| Weighted Average | 0.94 | 0.94 | 0.94 | 114 |

Table 3 offers a comprehensive view of the model's performance for each class (0 and 1), along with overall averages:

i. **For class 0 (negative cases):**

- Precision: 0.91
- Recall: 1.00
- F1-Score: 0.95
- Support: 71 cases

ii. **For class 1 (positive cases):**

- Precision: 1.00
- Recall: 0.84
- F1-Score: 0.91
- Support: 43 cases

The model excels at identifying negative cases (class 0), achieving perfect recall (1.00). This means it correctly identified all negative cases in the test set. For positive cases (class 1), the model achieves perfect precision (1.00), indicating that when it predicts a positive case, it is always correct. However, its recall for positive cases (0.84) is lower, suggesting it misses some positive cases. The F1-scores, which balance precision and recall, are high for both classes (0.95 for class 0 and 0.91 for class 1), indicating good overall performance.

The macro average (simple average across classes) and weighted average (average weighted by support) both show high values (0.93 and 0.94 respectively) for precision, recall, and F1-score. This suggests that the model maintains good balance in its performance across classes, even with the uneven distribution of cases (71 for class 0, 43 for class 1).

### *E. Limitations*

While the results are promising, it's important to consider the following limitations:

i. The model's perfect precision comes at the cost of lower recall (0.8372). In a clinical context, this means that while the model never misclassifies a benign tumour as malignant, it misses about 16.28% of malignant tumours. This trade-off needs careful consideration in real-world applications.
ii. Dataset of 569 cases was used in the model training. A larger, more diverse dataset would be necessary to ensure the model's generalizability across different patient populations and tumour types.
iii. The perfect precision on the train dataset, while impressive, raises questions about potential overfitting. Further validation on completely independent datasets would be valuable.
iv. The model relies on a specific set of features (size, shape, texture). While these align with clinical understanding, there may be other important factors not captured in this dataset.
v. Lack of Real-World Clinical Validation: The model has only been tested on the Kaggle dataset, not in actual clinical settings. Performance in real-world medical environments remains unverified.
vi. Absence of Prospective Testing: Lack of evaluation on completely new, unseen patient data in a controlled clinical trial setting.
vii. The current model's recall of 0.8372 requires significant improvement for clinical deployment. A multi-faceted approach is proposed to enhance detection rates whilst maintaining practical clinical utility.

### *F. Future Directions*

To address these critical limitations and enhance the robustness and clinical relevance of our model, we propose the following strategies:

1. Collaboration with Medical Institutions:
   - Establish partnerships with multiple healthcare providers across diverse geographical locations. These are a few specific medical institutions:
     - University teaching hospitals with diverse patient populations
     - Research-oriented clinical environments with established protocols for clinical studies
     - Access to multidisciplinary expertise and advanced technological infrastructure
     - Experience in conducting clinical trials and managing research protocols
   - Implement the model in real clinical environments under controlled conditions.
   - Conduct prospective studies to assess the model's performance in real-time clinical decision-making processes.
   - Gather feedback from healthcare professionals to refine the model's usability and integration into clinical workflows.

2. Cross-Validation on Multiple Diverse Datasets:

   - Extend the study to include rigorous cross-validation using multiple datasets from different regions, populations, and healthcare systems.
   - Analyze the model's performance across various demographic groups, including different age ranges, ethnicities, and socioeconomic backgrounds.
   - Assess the model's robustness in handling variability in data collection methods and quality across different clinical settings.

3. Longitudinal Performance Evaluation:

- Conduct long-term follow-up studies to assess the model's performance over time.
- Evaluate the model's ability to adapt to evolving clinical practices and potential shifts in disease patterns.

4. Comparative Analysis with Existing Clinical Methods:

- Perform head-to-head comparisons between our model and current gold standard diagnostic methods.
- Assess the model's potential to complement or enhance existing clinical decision-making processes.

5. Ethical and Regulatory Considerations:

- Engage with relevant regulatory bodies to ensure compliance with healthcare data protection and ethical guidelines.
- Develop protocols for responsible AI implementation in clinical settings, addressing issues of transparency, explainability, and accountability.

6. Probabilistic Threshold Adjustment
   - Develop an adaptive thresholding system that adjusts based on patient risk factors
   - Implement lower classification thresholds for high-risk populations
   - Create separate threshold levels for different cancer types based on their severity and progression rates

7. Cost-Sensitive Learning
   - Incorporate misclassification costs into the model training process
   - Weight false negatives (missed cancers) significantly higher than false positives
   - Adjust weights based on clinical outcome data and expert knowledge

8. Ensemble Approach
   - Combine multiple models with different threshold settings
   - Implement a voting system weighted towards positive predictions
   - Use specialist models for different patient subgroups
9. Independent Dataset Validation

A comprehensive independent validation strategy represents a critical component of our future work. This involves undertaking thorough validation studies using entirely independent datasets not involved in the model's development or initial testing phases. We will evaluate model performance across datasets collected from different medical equipment manufacturers and imaging protocols to assess vendor-agnostic capabilities.

Collaboration with international research institutions will be established to access diverse and independent validation cohorts, ensuring broad geographical and demographic representation. Standardised preprocessing pipelines will be implemented to ensure consistent data handling across validation datasets whilst maintaining their independent characteristics.

This comprehensive validation strategy aims to ensure the model's reliability, generalisability, and clinical utility across diverse healthcare settings and patient populations. Through rigorous validation and continuous assessment, we seek to establish our model as a robust and trustworthy tool for clinical application.

# IX. Model Expalanability

To enhance the transparency and interpretability of our machine learning model, we employed three distinct explainability techniques: SHapley Additive exPlanations (SHAP), Local Interpretable Model-agnostic Explanations (LIME), and Permutation Importance using the ELI5 library. Each method provides unique insights into the model's decision-making process, contributing to a comprehensive understanding of its behaviour.

## *A. SHAP-Deep Explainer*

The Deep Explainer, tailored for neural network architectures, computed the average contribution of each feature to the model's output magnitude. The values measure the average contribution of each feature to the model output magnitude. A higher value means a higher relevance, while a lower SHAP value means a lower relevance.

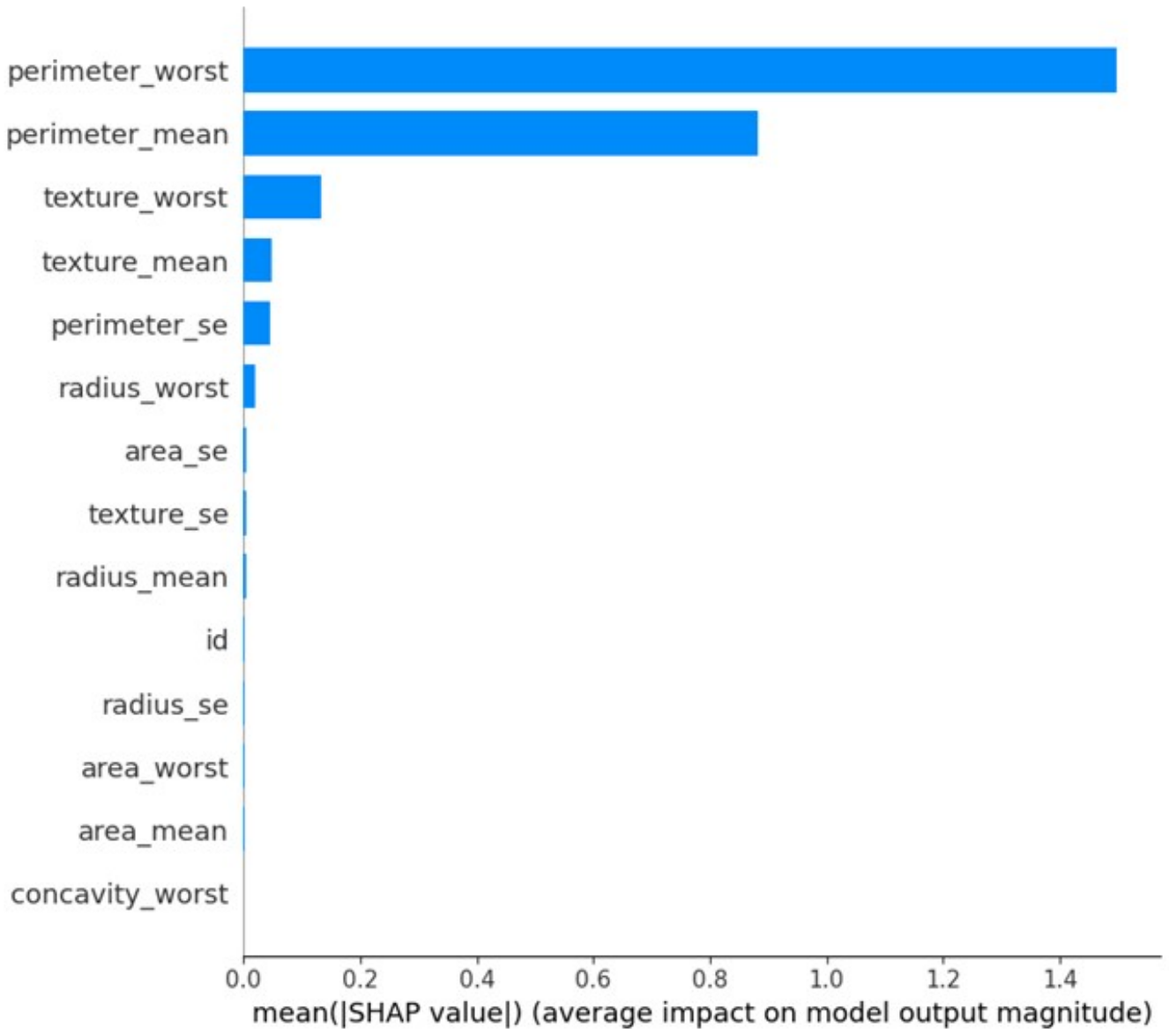

Fig. 17. Bar-Chart of SHAP Values for Deep Explainer.

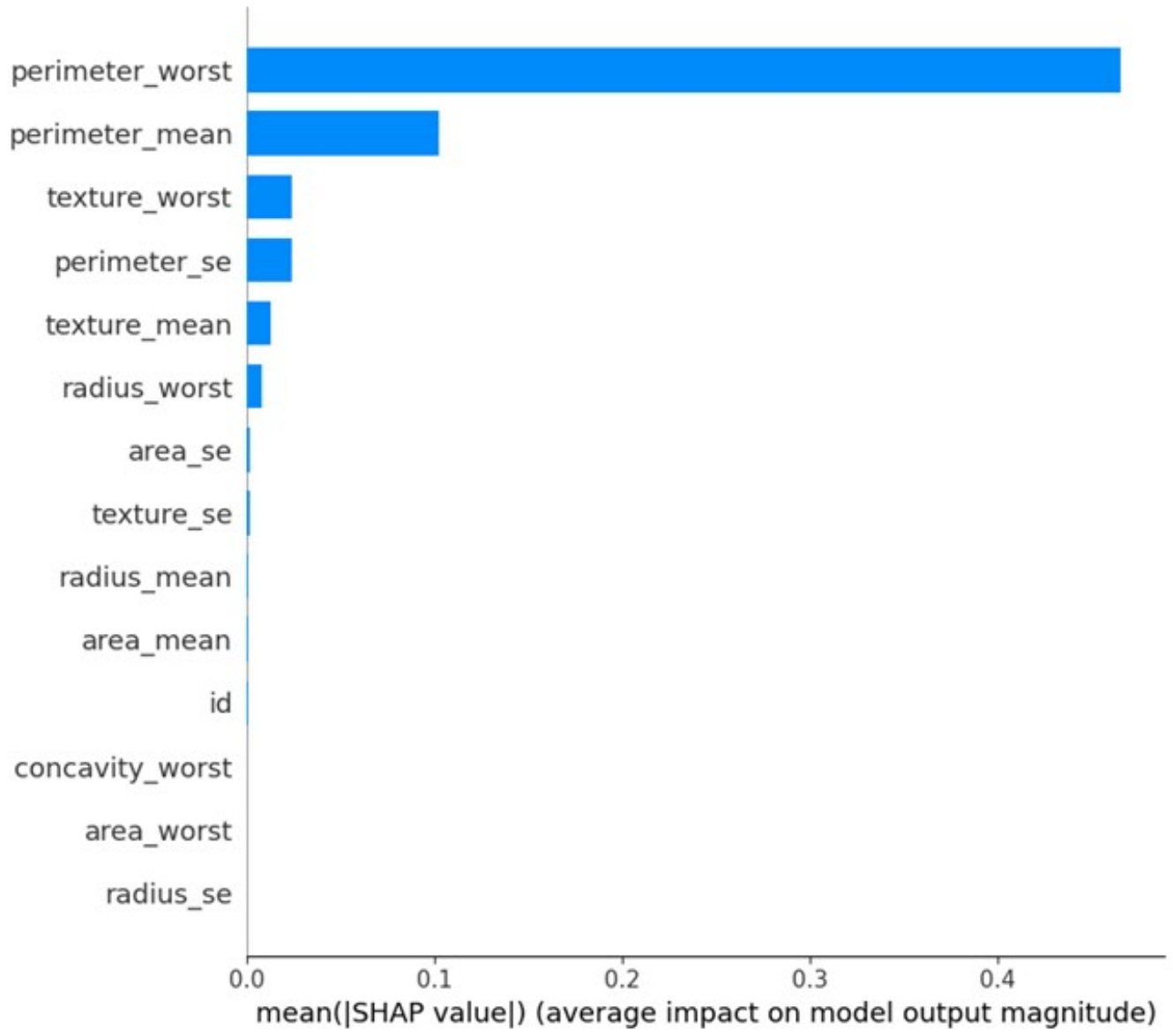

Fig. 18. Bar-Chart of SHAP Values for Kernel Explainer.

Fig. 11 illustrates the SHAP (SHapley Additive exPlanations) DeepExplainer values for various features of the dataset. The features are displayed on the y-axis, comprising attributes such as perimeter, texture, radius, area, and concavity in their worst, mean, or standard error (se) measurements. The x-axis represents the mean SHAP value, ranging from 0 to approximately 1.4. The 'perimeter_worst' feature exhibits the highest mean SHAP value, indicating that it has the most substantial average influence on the model's output magnitude. This is followed by 'perimeter_mean', which also shows a significant impact. 'texture_worst' and 'texture_mean' demonstrate notable influences, albeit to a lesser degree than the perimeter measurements. 'radius_worst' rounds out the top five most influential features.

Interestingly, some features such as 'radius_se', 'area_worst', 'area_mean', and those below them on the chart, display very low or negligible mean SHAP values, suggesting they have minimal impact on the model's predictions.

This bar plot reveals that the model is more responsive to perimeter and texture measurements compared to radius and area measurements. Additionally, it indicates that the 'worst' and 'mean' measurements generally carry more weight than the standard error (se) measurements in the model's decision-making process. It's worth noting that only a subset of the dataset's features is shown in this plot, focusing on those with the highest mean SHAP values. Features not displayed likely have even lower influence on the model's output.

### *B. SHAP–Kernel Explainer*

To validate our findings, we employed the model-agnostic Kernel Explainer. As shown in Figure 12, this method corroborated the results obtained from the Deep Explainer, reinforcing the primacy of perimeter and texture measurements in the model's decision-making process.

Fig. 12 shows the SHAP (SHapley Additive exPlanations) values from a Kernel Explainer. The features are shown on the y-axis, and they are things like perimeter, texture, radius, area, and concavity in their worst, mean or standard error (se) measurements. The x-axis shows the mean SHAP value, which changes from 0 to about 1.4. The “perimeter worst” feature has the highest mean SHAP value, which means that it has the biggest average effect on the model output size. Other features like “perimeter mean,“ “texture worst,“ and “texture mean” also have big effects but not as much, while features like “radius se,” “area worst,” and “area mean” have small effects. The bar plot shows that the model cares more about the perimeter and texture measurements than the radius and area measurements. It also shows that the worst and mean measurements are more important than the standard error measurements. This bar plot is similar to the one from the Deep Explainer, which also shows that the “perimeter worst” feature is the most important feature for the model prediction.

### *C. LIME Explainable Framework*

LIME was employed to elucidate individual predictions by approximating the model locally with an interpretable surrogate. Figure 13 presents a visual report generated by LIME, detailing prediction probabilities and feature contributions for specific cases. This analysis reveals how individual feature values influence predictions, either positively or negatively. The LIME results demonstrate a consistent alignment between the model's predictions and the corresponding feature values, further validating the model's coherence.

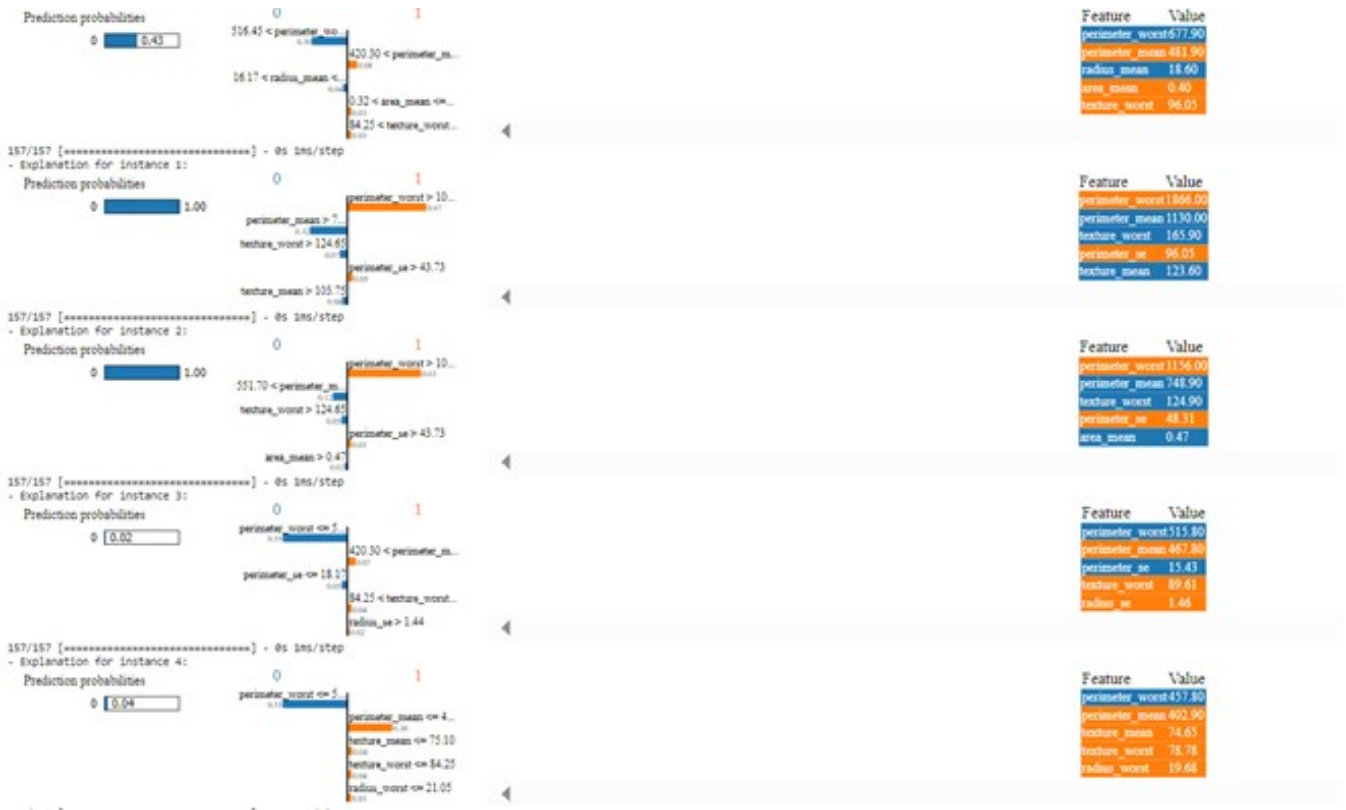

Fig. 19. Visualization of LIME Explainer.

Fig. 13 shows how the LIME (Local Interpretable Model-agnostic Explanations) explainable framework helps us understand the predictions of complex machine learning models. The image shows the prediction probabilities and the feature contributions to those predictions for each case. The prediction probabilities show how sure the model is about its predictions, while the feature contributions show how much each feature affects the prediction in a positive or negative way. The features are things like perimeter, texture, radius, area, and concavity in their worst, mean or standard error (se) measurements. The values of the features are marked in orange, with different shades showing the level of influence on the prediction. The image also shows that the model cares more about the perimeter and texture measurements than the radius and area measurements. We can also say that the worst and mean measurements are more important than the standard error measurements. The image also shows that the model's predictions match the feature values, as the features with higher values tend to have positive contributions and the features with lower values tend to have negative contributions. For example, the first case has a high value for perimeter worst and a low value for texture worst, and the model predicts it as class M "Malignant" with a high probability. The second case has a low value for perimeter worst and a high value for texture worst, and the model predicts it as class B "Benign" with a high probability.

### D. Eli5 – Permutation Importance

To complement our feature importance analysis, we utilised the permutation importance module from the ELI5 library. This technique assesses feature importance by measuring the impact of random feature value permutations on overall model performance. As depicted in Figure 14, 'area worst' emerged as the most critical feature, followed by 'area se' and 'radius se'. This analysis suggests that area and perimeter measurements exert a more profound influence on model predictions compared to radius and texture measurements. Furthermore, it indicates that worst and standard error (se) measurements generally carry more weight than mean measurements in the model's decision-making process.

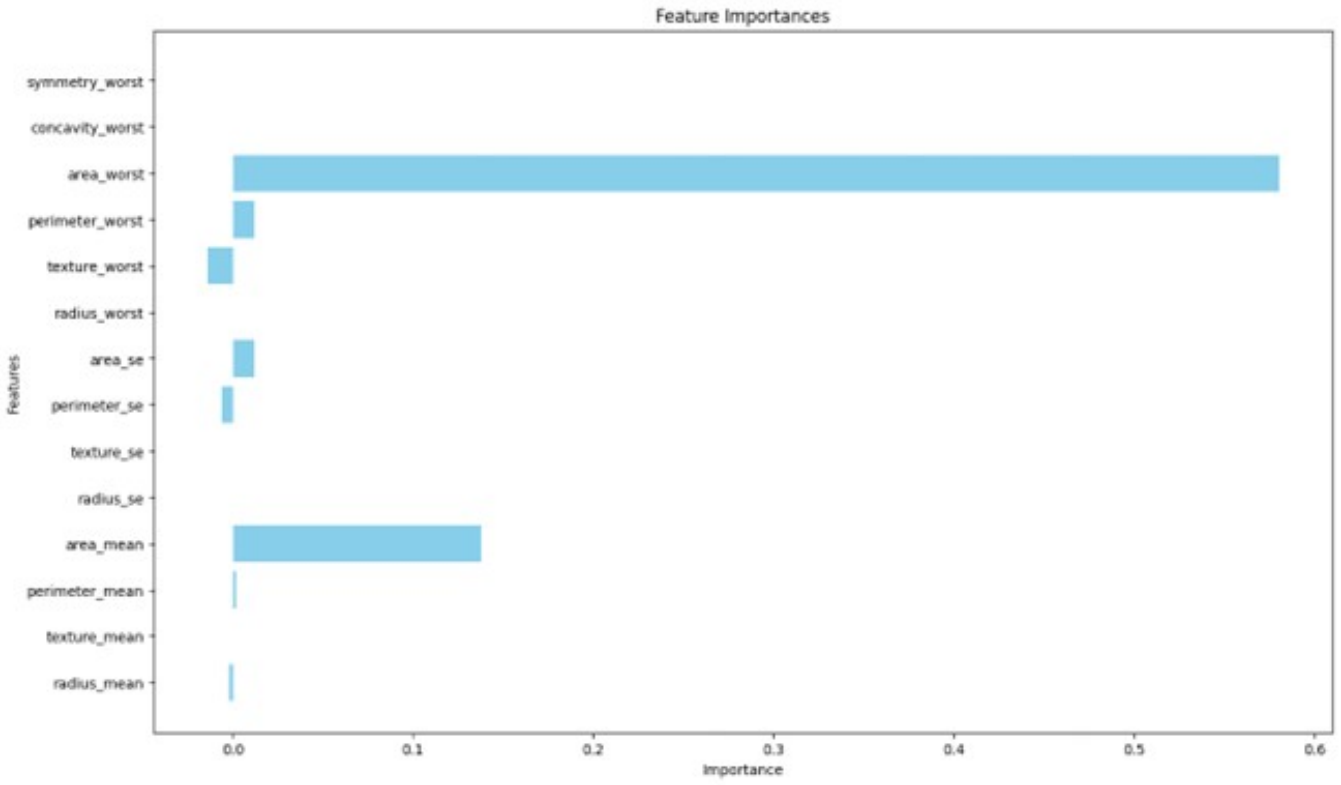

Fig. 20. Bat-Chart of Feature Importance.

Fig. 14 is a horizontal bar plot that shows the feature importances from the "get score importance" module of eli5.permutation importance. This module calculates the importance of each feature by randomly changing its values and seeing how the model's score changes. A bigger change means a more important feature, while a smaller change means a less important feature. The features are shown on the left side of the image, and they are things like perimeter, texture, radius, area, and concavity in their worst, mean or standard error (se) measurements. The right side of the image shows the importance, which changes from 0.0 to 0.6. The "area worst" feature has the highest importance, which means that it has the biggest impact on the model's score. Features like "area se," "radius se," and "perimeter se" also have big impacts but not as much, while features like "symmetry worst" and "concavity worst" have small impacts. The image shows that the model cares more about the area and perimeter measurements than the radius and texture measurements. It also shows that the worst and se measurements are more important than the mean measurements.

***Clinical Decision Support Process using XAI Methods***

1. Initial Assessment:
   - Review raw clinical images and traditional metrics
   - Generate ML model prediction

2. Explainability Analysis:
   - Run SHAP analysis for feature importance
   - Generate LIME explanation for specific case
   - Compare against known malignancy patterns

3. Clinical Synthesis:
   - Combine ML insights with clinical expertise
   - Document key supporting and contradicting evidence
   - Prepare patient-friendly explanation

## X. Conclusion and Limitation

### A. Conclusion

This research endeavour represents a big stride in improving cancer detection and prognosis by establishing an advanced AI

model. The global effect of cancer, highlighted by millions of deaths annually, underscores the demand for creative remedies that surpass existing methods. The World Health Organization's prediction of roughly 10 million cancer-related deaths in 2020 highlights the necessity for early detection and exact diagnosis, creating the cornerstone of this research. The suggested AI approach includes powerful deep learning techniques, leveraging neural networks to examine big datasets and detect relevant patterns. Recognizing AI's potential to improve healthcare, especially in cancer care, this initiative tackles difficulties associated with existing diagnostic and treatment approaches, such as high prices, low accuracy, and lengthy procedures with unpleasant side effects. A fundamental contribution of this research is its focus on explainable AI (XAI) methodologies to enhance the transparency and interpretability of the model. The worry regarding the "black box" character of many AI models, particularly those based on deep learning, is addressed by adopting cutting-edge XAI frameworks and modules. This technique tries to demystify the AI model's decision-making processes, providing explicit insights into the elements influencing forecasts and increasing user trust. The research project's comprehensive approach encompasses system design, architecture, and implementation, ensuring a full and attentive developmental process. Initial procedures, such as selecting and curating appropriate datasets, rigorous data pre-processing, and feature engineering, highlight privacy, security, and fairness issues. The installation of a deep learning model, illustrated by the Convolutional Neural Network diagram, represents the technological foundation driving the AI system's predictive capabilities. Moreover, the initiative understands the crucial relevance of data visualization in efficiently presenting results to users. Visual representations, including charts, histograms, and heatmaps, are vital in presenting insights from datasets and providing a clear grasp of the model's predictions and interpretability. The study project's goals, summarized in a series of questions and accompanying objectives, reflect a dedication to enhancing AI-assisted cancer detection. The constant quest of accuracy, explainability, fairness, ethics, and trustworthiness are at the centre of the research agenda. Systematic testing and verification processes strive to assess the model's performance against these criteria, ensuring it satisfies the highest standards of reliability and effectiveness. In summary, this research initiative intends to transcend present constraints in cancer detection and prediction approaches. By establishing an AI model that produces accurate results and encourages understanding and trust, the project envisions a future where AI-assisted clinical decision-making becomes a vital and ethically sound instrument in combating cancer. Through thorough research, innovative technologies, and a strong dedication to transparency, this project significantly contributes to the ongoing evolution of healthcare, marking a key step towards a more educated, accessible, and compassionate approach to cancer care.

### *B. Limitations*

- Limitations of the tabular data: The cancer data is a tabular dataset that has numerical features, such as radius, texture, perimeter, area, and concavity of the cells. However, this type of data has some limitations, such as:
    - Lack of spatial information: The tabular data does not capture the spatial relationships or patterns of the cells, which may be relevant for cancer detection. For example, the form, size, and orientation of the nucleus may reflect the malignancy of the cells, but these are not shown in the tabular data.
    - Need for feature engineering: The tabular data requires a lot of pre-processing and feature engineering before feeding the model to produce predictions. For example, the data needs to be standardized, normalized, imputed, encoded, and selected. This can add errors, biases, or noise in the data, or limit the interpretability of the model.
- Precision-recall trade-off: While the model achieved perfect precision (1.0000), it came at the cost of lower recall (0.8372). This means that while the model never misclassifies a benign tumor as malignant, it misses about 16.28% of malignant tumors. In a clinical context, this trade-off needs careful consideration as missing malignant cases could have serious consequences.

### *C. Ethical Considerations*

The application of artificial intelligence in medical diagnosis, particularly in life-or-death situations such as cancer detection, necessitates careful consideration of ethical implications. This section discusses potential biases, the importance of fairness, and the need for accountability in AI-assisted medical decision-making systems.

1. Potential Biases in the AI System

- Data Representation Bias: Our model was trained on a specific dataset that may not fully represent the diverse population it could potentially serve. This could lead to varying performance across different demographic groups.
- Feature Selection Bias: The tabular data used in our study relies on pre-selected features. There's a risk that important indicators of cancer that are not represented in these features could be overlooked.
- Algorithmic Bias: The machine learning algorithms themselves may have inherent biases that could affect their decision-making processes, potentially leading to unfair outcomes for certain groups.

2. Importance of Fairness

- Equal Access to Healthcare: AI systems should provide equally accurate diagnoses regardless of a patient's demographic characteristics.

- Trust in Healthcare Systems: Fairness in AI systems is essential for maintaining public trust in healthcare institutions that adopt these technologies.

- Legal and Ethical Obligations: Healthcare providers have a moral and often legal obligation to provide fair and unbiased care to all patients.

3. Accountability in AI-Assisted Medical Decisions

- Human Oversight: While AI can assist in diagnosis, final decisions should involve human medical professionals who can interpret AI outputs in the context of broader patient information.

- Explainable AI: Efforts should be made to use interpretable AI models that can provide clear reasoning for their diagnoses, allowing for scrutiny and validation by medical professionals.

- Regular Audits: Systematic audits of the AI system's performance across different patient groups should be conducted to identify and address any emerging biases or inconsistencies.

4. Frameworks for Ensuring Ethical Use

- Continuous Monitoring and Bias Detection:
  - Implement automated systems to continuously monitor the AI's performance across different demographic groups.
  - Utilize statistical methods to detect any significant disparities in diagnostic accuracy among subpopulations.

- Diverse and Representative Data Collection:
  - Regularly update and diversify the training data to ensure it represents the population the system serves.
  - Collaborate with multiple healthcare institutions to gather a more comprehensive and diverse dataset.

- Algorithmic Fairness Techniques:
  - Employ pre-processing techniques to balance the training data across different groups.
  - Utilize in-processing methods that enforce fairness constraints during model training.
  - Apply post-processing techniques to adjust model outputs for improved fairness.

- Ongoing Education and Training:
  - Provide regular training to healthcare professionals on the capabilities and limitations of the AI system.
  - Educate patients on the role of AI in their diagnosis and their rights regarding AI-assisted decisions.

`

**Badaru I. Olumuyiwa** received the M.Sc. degree in electrical and electronics engineering (communication option) from the University of Lagos, Lagos, Nigeria, and the M.Sc. degree in artificial intelligence from Teesside University, Middlesbrough, UK, in 2023. His major field of study is artificial intelligence with a focus on explainable AI in healthcare.

He has worked in the telecommunications sector, gaining valuable experience in technology applications. His current research interests include explainable AI, deep learning techniques, and their applications in medical diagnostics, particularly cancer diagnosis and prediction.

Mr. Olumuyiwa recently completed his master's thesis titled "A Trustworthy Explainable AI Model: Cancer Diagnosis and Prediction with Deep Learning Techniques," which explores the potential of explainable AI models to improve diagnostic accuracy and foster trust in healthcare settings. His research aims to address the critical need for transparent and interpretable AI solutions in medical diagnostics.